%% file: main_with_supp.tex
\documentclass[lettersize,journal]{IEEEtran}
\usepackage{amssymb}
\usepackage[utf8]{inputenc} %
\usepackage[T1]{fontenc} %
\usepackage{url} %
\usepackage{booktabs} %
\usepackage{amsfonts} %
\usepackage{amsmath}
\usepackage{amssymb}
\usepackage{amsthm}
\usepackage{nicefrac} %
\usepackage{microtype} %
\usepackage{doi}
\usepackage[caption=false, font=footnotesize]{subfig}
\usepackage[dvipsnames, table]{xcolor} %
\usepackage{multirow}
\usepackage{floatrow}
\usepackage{blindtext}
\usepackage{bm} %
\usepackage{upgreek} %
\usepackage{listings} %
\usepackage[inline]{enumitem} %
\setlist{nosep} %
\usepackage{tikz}
\usepackage{xcolor}
\usepackage{balance}
\usepackage{amsthm}%
\usepackage{mymacros}
\usepackage{soul}
\usetikzlibrary{patterns} %
\usepackage{balance} %
\usepackage{dblfloatfix}
\usepackage{cite} 
\usepackage{makecell}
\usepackage[symbol*]{footmisc}
\usepackage{tabularx}
\usepackage{tablefootnote}
\usepackage{wrapfig}
\usepackage{floatrow} %
\hypersetup{
    colorlinks,
    linkcolor={black},
    citecolor={black},
    urlcolor={blue}
}

\graphicspath{{figures/}}

\makeatletter
\def\input@path{{figures/}}
\makeatother

\newcolumntype{b}{X}
\newcommand{\heading}[1]{\multicolumn{1}{c}{#1}}

\newtheorem{claim}{Claim}

\begin{document}

\title{Scalable and Efficient Continual Learning from Demonstration via a Hypernetwork-generated Stable Dynamics Model}

\author{
  Sayantan Auddy\IEEEauthorrefmark{1},
  Jakob Hollenstein,
  Matteo Saveriano,
  Antonio Rodr\'{i}guez-S\'{a}nchez and 
  Justus Piater
  \thanks{Sayantan Auddy is with the Faculty of Electrical Engineering and Computer Science, Technical University of Berlin, Germany (email: auddy@tu-berlin.de). \IEEEauthorrefmark{1}Corresponding author.
  }
  \thanks{Jakob Hollenstein and Justus Piater are with the Department of Computer Science, University of Innsbruck, Austria (email: \{\mbox{firstname.lastname}\}@uibk.ac.at). Justus Piater is also with the Digital Science Center (DiSC), University of Innsbruck, Austria.} 
  \thanks{Matteo Saveriano is with the Department of Industrial Engineering, University of Trento, Italy (email: matteo.saveriano@unitn.it).}
  \thanks{Antonio Rodr\'{i}guez-S\'{a}nchez is with the Singular Research Center on Intelligent Systems (CiTIUS), University of Santiago de Compostela, Spain (email: antoniojose.rodriguez@usc.es).}
  \thanks{Sayantan Auddy was supported by a doctoral scholarship granted by the University of Innsbruck, Vice-Rectorate for Research. This work was also funded by the European Union project INVERSE (GA no. 101136067).}
  }

\maketitle
  
\begin{abstract}
Robots capable of learning from demonstration (LfD) must exhibit stability while executing learned motion skills. To be effective in the real world, they should also remember multiple skills over time -- a capability lacking in current stable-LfD methods. 
We propose an approach to stable, continual LfD, and highlight the role of stability in improving continual learning. Our proposed hypernetwork generates the parameters of two neural networks: a trajectory learning dynamics model, and a trajectory-stabilizing Lyapunov function.
These generated networks form a clock-augmented stable neural ODE solver (sNODE), a stable dynamics model that offers a superior stability-accuracy trade-off compared to the state-of-the-art.
We further propose stochastic hypernetwork regularization with a single, uniformly-sampled task embedding, reducing the cumulative training time for $N$ tasks from O($N^2$) to O($N$) without degrading performance on real-world tasks.
We introduce high-dimensional variants of the popular LASA dataset to assess scalability and extend a dataset of robotic LfD tasks to assess real-world performance.
We empirically evaluate our approach on multiple LfD datasets of varying complexity, including sequences of 7--26 tasks, trajectories of 2--32 dimensions, and real-world tasks involving position and orientation.
Our thorough evaluation on multiple LfD datasets demonstrates that our approach sequentially learns and retains multiple motion skills without retraining on past demonstrations, 
and outperforms other relevant baselines in terms of trajectory errors, continual learning scores, and stability metrics.
Notably, we show that stability greatly enhances continual learning performance, particularly in size-efficient chunked hypernetworks. Our code is available at \url{https://github.com/sayantanauddy/clfd-snode}.

\end{abstract}

\begin{IEEEkeywords}
Learning from Demonstration, Continual Learning, Hypernetworks
\end{IEEEkeywords}

\section{Introduction}
\label{sec:intro}

\begin{figure*}[t]
  \centering
  \includegraphics[width=0.7\textwidth]{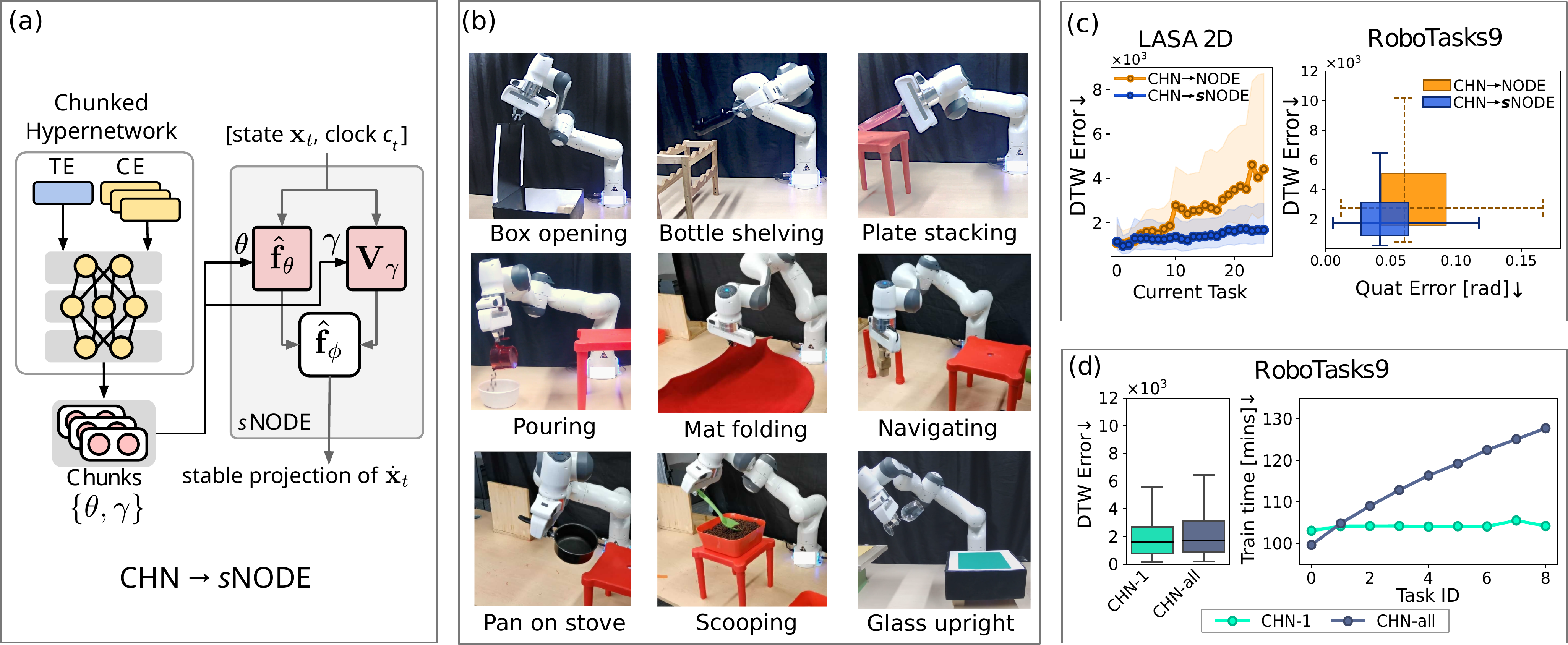}
  \caption{
  \newer{
  Overview of our proposed approach and main results.
  ~\emph{(a)}~\chnsnode{}: a chunked hypernetwork (CHN) accepts a task embedding (TE) and a set of chunk embeddings (CE) as input and generates parameters $\upphi=\{\uptheta, \upgamma\}$ of our proposed clock-augmented stable NODE (\snode{}) $\hat{{\mathbf{f}}}_\upphi$, comprising a nominal dynamics model $\hat{\mathbf{f}}_\uptheta$ and a Lyapunov function $V_\upgamma$. 
  Colors differentiate task-specific parameters~\legendsquare{myblue}, regularized (task-independent) parameters~\legendsquare{myyellow}, and non-trainable outputs~\legendsquare{mypink}~(details in \mysection{sec:methods});
  ~\emph{(b)}~The nine real-world LfD tasks of RoboTasks9. The last 5 tasks are introduced in this paper (details in \mysection{sec:experiment_setup}). Our proposed approach learns all tasks continually with a single hypernetwork;%
  ~\emph{(c)}~Stability elevates CL performance -- \chnsnode{} outperforms NODE-based solution~(details in \mysection{sec:experiments_results});
  ~\emph{(d)}~Stochastic regularization with a single task embedding (CHN-1) reduces training cost of $N$ tasks from $\mathcal{O}(N^2)$ to $\mathcal{O}(N)$ but performs as well as full regularization (CHN-all) on real-world tasks (details in \mysection{sec:experiments_results}).
  }
  }
  \label{fig:intro}
\end{figure*}

\emph{Learning from demonstration} (LfD) is a natural way for humans to impart movement skills to robots, particularly when the desired motion cannot be hardcoded or defined in terms of an optimization objective, but can be easily demonstrated~\cite{ravichandar2020recent}. 
Methods for LfD are required not only to reproduce the motion demonstrated by humans, but also to guarantee the stability of the produced motion. Stable motion implies that the robot's trajectory does not diverge and it does not assume unsafe configurations during its motion. 
Learning stable motion trajectories has been the focus of several LfD techniques \cite{hersch2008dynamical, saveriano2020energy, khansari2011learning, khansari2014learning, ijspeert2013dynamical, saveriano2023dynamic}. These approaches encode the demonstrations as stable dynamical systems, and are the state-of-the-art in LfD~\cite{ravichandar2020recent}.
However, a limitation of these approaches lies in their reduced representational capabilities, especially in high-dimensional spaces~\cite{sochopoulos2024learning, moerland2023model}. 
This makes it important to develop neural network-based LfD techniques that can utilize high-dimensional features~\cite{sochopoulos2024learning, rahmatizadeh2018vision, byravan2017se3}, as also proposed in this paper.

A number of recent works on stable dynamical systems utilize neural networks for learning from observations~\cite{urain2020imitationflow, kolter2019learning, sochopoulos2024learning, lawrence2020almost, richards2018lyapunov}. 
However, these methods focus on learning only a single \emph{skill}. To learn a new skill, the model must be trained from scratch on new demonstrations and the previously learned skill is forgotten as the network parameters are optimized on the new task. An LfD-capable robot in the real world should ideally be capable of \emph{continual learning}~(CL)~\cite{parisi2019CL_survey}, i.e., it should learn new skills as and when needed, and also retain past knowledge. 
Recent work on \emph{continual}~LfD~\cite{auddy2022continual} has shown that a single \emph{hypernetwork}~\cite{ha2017hypernetworks, von2019continual} that generates \emph{neural ordinary differential equation solvers}~(NODEs)~\cite{chen2018neural}, can continually learn and remember a sequence of motion skills.
However, this approach has limitations such as degrading performance on long task sequences, lack of stability guarantees \new{about the convergence of the predicted motion to a goal},
and a linear increase in training time for each new task.

To overcome these limitations, we propose a continual LfD system comprising a single hypernetwork that generates the parameters of two neural networks constituting a \emph{stable} NODE (\snode{}): a network representing a nominal dynamics model, and a parameterized Lyapunov function for ensuring stability~\cite{kolter2019learning}. 
The hypernetwork forms the \emph{continual learning mechanism} responsible for retaining knowledge from previous demonstrations and the \snode{} represents the \emph{task-learner} responsible for learning the current LfD task. 
The entire system is trained end-to-end via supervised learning using only the demonstrations for the current LfD task and does not need to store or retrain on data from past tasks.
For the task-learner, we introduce a state-independent, monotonic and bounded clock signal to an existing \emph{stable} NODE (\snode{}) architecture~\cite{kolter2019learning} to improve accuracy while retaining stability.
We also propose a stochastic hypernetwork regularization strategy to improve training efficiency.

\new{
We incorporate motion stability with the \snode{}, which inherently approaches an equilibrium point due to its architectural inductive biases~\cite{kolter2019learning}. Though motion stability is not directly related to continual learning stability (avoidance of catastrophic forgetting), we empirically show that the \snode{}'s hardwired ability of converging at the equilibrium point bolsters the continual learning performance of the overall system.
}

We perform experiments on the popular LASA benchmark~\cite{khansari2011learning} and the \emph{HelloWorld} dataset~\cite{auddy2022continual} of complex \mbox{2-dimensional} trajectories. 
To assess scalability to higher dimensions, we combine multiple tasks from the original LASA dataset~\cite{khansari2011learning} to create new datasets of \mbox{8-},\mbox{16-}, and \mbox{32-dimensional} trajectories and use them in our evaluations.
We add 5 new tasks to the \mbox{\emph{RoboTasks}} dataset~\cite{auddy2022continual} to create \emph{\robotasks{}}, a dataset of 9 real-world LfD tasks that is also used for evaluation. 
We report quantitative metrics for predicted trajectories, CL performance, and stability. We also perform qualitative evaluations with a physical Franka Emika Panda robot.
Our 
evaluations show that the stable nature of our continual LfD system elevates CL performance 
and scales effectively to long task sequences and high-dimensional trajectories. 
We also show that our hypernetwork-based approach empirically outperforms other CL mechanisms~\cite{parisi2019CL_survey,wang2024comprehensive}.
Our stochastic regularization technique for hypernetworks reduces the cumulative training time of $N$ tasks from $\mathcal{O}(N^2)$ to $\mathcal{O}(N)$, without impacting performance in real-world tasks.
\myfigure{fig:intro} presents an overview of our approach and some key results. Our open-source code and proposed datasets %
will be released upon acceptance of this paper.
In summary, our main contributions are:
\begin{itemize}
  \setlength\itemsep{0pt}
  \item We propose a  \emph{stable and continual} LfD approach that utilizes a single hypernetwork$\rightarrow$\snode{} model to continually learn multiple LfD tasks, showing that stability improves continual learning performance.
  \item We introduce a stochastic hypernetwork regularization technique that reduces the cumulative training cost for $N$ tasks from $\mathcal{O}(N^2)$ to $\mathcal{O}(N)$ without loss of performance on real-world tasks.
  \item We create high-dimensional versions of the LASA dataset, and add 5 new tasks to the RoboTasks dataset~\cite{auddy2022continual}, forming \emph{\robotasks{}}, a dataset of 9 real-world LfD tasks.
\end{itemize}

\section{Related Work}
\label{sec:related}

\new{
\noindent The work in this paper builds on existing techniques in LfD~\cite{kolter2019learning} and CL~\cite{von2019continual}, and hence we review relevant literature from both fields. We also discuss methods that, like ours, address continual LfD. Whenever applicable, we highlight the differences between our approach and existing ones.
}
\smallskip

\emph{Learning from demonstration} (LfD)
is a valuable technique for teaching robots tasks that are difficult to program explicitly or learn from scratch, allowing them to bootstrap from human expertise and adapt to various situations in complex domains~\cite{ravichandar2020recent}.
Demonstrations used for training may be provided by different means~\cite{ravichandar2020recent}, including kinesthetic teaching~\cite{argall2009survey, billard2016learning, ravichandar2020recent, ahmadzadeh2018trajectory}, tele-operation~\cite{abbeel2010autonomous}, or passive observation~\cite{gao2021cril}.
Different learning approaches, including supervised learning~\cite{wu2010towards, argall2011teacher}, constrained optimization~\cite{englert2017inverse}, reinforcement learning~(RL)~\cite{calinon2013compliant}, and inverse RL~\cite{das2021model} have been utilized for LfD. In addition to Euclidean space, LfD in non-Euclidean spaces such as Riemannian manifolds is also a topic of current research~\cite{saveriano2023learning, urain2023se}.
\new{In this paper,} we focus on \emph{trajectory-based} learning, which is a sub-field of LfD~\cite{ravichandar2020recent}, \new{and utilize kinesthetic teaching for providing demonstrations to the robot (though trajectory data collected via other means would also suffice).}
Trajectory-based methods either fit probability distributions to the observed data with \emph{generative models}~\cite{hersch2008dynamical, khansari2011learning, urain2020imitationflow, figueroa2018physically}, or fit a discriminative model to the training data using function approximators such as neural networks~\cite{kolter2019learning, Ijspeert2002Movement}. 
The observed demonstrations can either be used to learn a static-mapping between time and the desired state of the robot, or to dynamically map the current robot state to the desired velocity~\cite{ravichandar2020recent}.

Different approaches ensure that the robot's motion is stable and convergent~\cite{hersch2008dynamical, khansari2014learning, saveriano2020energy}. 
Recent contributions in this area also include methods based on neural networks~\cite{kolter2019learning, sochopoulos2024learning, lawrence2020almost, richards2018lyapunov, urain2020imitationflow, ravichandar2020recent}.
These neural network-based methods are attractive as they can be scaled to utilize high-dimensional features~\cite{sochopoulos2024learning, moerland2023model}.
While some methods rely on normalizing flows~\cite{urain2020imitationflow}, others enforce stability with a learned Lyapunov function~\cite{sochopoulos2024learning, kolter2019learning, lawrence2020almost, richards2018lyapunov}.
\new{
Due to the long training times of normalizing flow-based LfD~\cite{urain2020imitationflow} we opt for a Lyapunov-based approach~\cite{auddy2022continual}.}
Among the Lyapunov-based approaches, some attempt to enforce stability with an extra training loss~\cite{richards2018lyapunov,chow2018lyapunov}. However, stability in unseen states may be difficult to prove~\cite{kolter2019learning}. Stability verification of a pre-trained dynamics model is also difficult~\cite{lawrence2020almost}. 
\notnew{
Hence, in this paper, we follow the approach of \emph{jointly} learning a dynamics model and a parameterized Lyapunov function~\cite{kolter2019learning,lawrence2020almost}.
}

\notnew{
Despite the maturity of research in trajectory-based LfD, the predominant focus is on acquiring a single skill~\cite{urain2020imitationflow,kolter2019learning}, which necessitates training a new model for each new skill. 
In contrast, our focus is on continually learning a sequence of LfD tasks with a single model.
Though multiple tasks can be learned with separate networks, a single continual LfD model is preferable as it eliminates the need to store a large number of networks for performing different tasks.
NODEs~\cite{chen2018neural} generated by a single hypernetwork have been used previously to learn multiple LfD skills continually~\cite{auddy2022continual}. However, NODEs lack stability guarantees, which may result in divergent robot motion.
In this paper, we propose \emph{stable, continual LfD}. 
We augment the dynamics model of~\cite{kolter2019learning} with a state-independent, monotonic and bounded clock signal that improves accuracy while retaining stability, and generate the two constituent networks of this model with a hypernetwork.
Our present approach learns multiple tasks and predicts stable motion. Crucially, the introduction of stability elevates CL performance, as we discuss later.}
\smallskip

\noindent\emph{Continual Learning} (CL) is a promising way for embodied agents like robots to gradually assimilate knowledge without needing to be preprogrammed for future tasks \emph{a priori}. 
Popular CL strategies~\cite{parisi2019CL_survey,delange2021continual,mundt2023wholistic,wang2024comprehensive} include parameter growth~\cite{rusu2016progressive}, replay of real~\cite{rebuffi2017icarl} or generated exemplars~\cite{shin2017continual} from past tasks, and regularization with additional constraints~\cite{kirkpatrick2017overcoming, zenke2017SI,aljundi2018MAS}.

Though most CL approaches are evaluated on image classification~\cite{lesort_continual_2020}, 
\new{
a few methods, including our own, apply CL in a robotics context.
}
In~\cite{thrun_lifelong_1995}, CL is applied to navigation and find-and-fetch tasks.
More recently, CL is used to adapt the perception and behavior models of social robots to changing human behavior~\cite{churamani_continual_2020, churamani_continual_2022}. 
A replay-based technique~\cite{lopez2017gradient} is used to continually learn navigation tasks in~\cite{liu_lifelong_2021}.
Regularization-based CL~\cite{kirkpatrick2017overcoming, zenke2017SI} is used to adapt the dynamics model of an industrial robot to changing conditions~\cite{trinh_development_2022}.
In~\cite{sarabakha_online_2023}, several CL methods~\cite{chaudhry2019continual,li2017learning,chaudhry2018efficient,kirkpatrick2017overcoming, chaudhry2018riemannian} are evaluated on mobile ground and aerial robots.
In~\cite{gao2021cril}, pseudo-training data of past tasks is created with generative replay~\cite{shin2017continual} to train a robot through imitation learning. 
Self-supervised task inference is used in a continual multitask learning setup in~\cite{hafez2023continual}.
In~\cite{josifovski_auddy2024continual}, regularization-based CL~\cite{kirkpatrick2017overcoming,schwarz2018progress} is combined with domain randomization~\cite{tobin2017domain} to achieve sim-to-real transfer.%
\new{
}

\new{

Recently, some approaches to continual LfD/imitation learning have been proposed. LOTUS~\cite{wan2024lotus} utilizes unsupervised skill discovery to construct a library of reusable sensorimotor skills (i.e. skill networks). It continually updates existing skills to prevent forgetting and incorporates new skills to address novel tasks. A meta-controller is also trained to combine the learned skills.
NBAgent~\cite{liang2024never} continually learns manipulation skills by leveraging language instructions and 3D visual information. It separates shared and task-specific knowledge using a modular approach, allowing the robot to learn new skills without forgetting old ones.
M2Distill~\cite{roy2024m2distill} is a distillation-based multi-modal approach to continual LfD that mitigates distribution shifts that typically lead to forgetting by preserving consistent latent representations across vision, language, and action modalities.
TAIL~\cite{liu2023tail} is also a multi-modal approach that utilizes a pretrained and frozen transformer policy
and task-specific adapters to learn new tasks while preserving past knowledge. The task-specific adapter modules are incorporated into the pretrained module using techniques commonly employed for large language models.
CRIL~\cite{gao2021cril} uses deep generative replay~\cite{shin2017continual} of video demonstrations and action-conditioned video prediction to reconstruct past state-action trajectories. The authors, however, highlight challenges in maintaining high-quality video generation over long task sequences.
Currently, we do not use vision or language modalities like~\cite{liang2024never, roy2024m2distill, liu2023tail}, and though our hypernetwork-based setup can be extended to incorporate vision and language inputs, we focus on learning using numerical robot states with comparatively smaller networks.
Unlike~LOTUS~\cite{wan2024lotus} and TAIL~\cite{liu2023tail}, we do not rely on a pretraining phase where many tasks are learned with multi-task learning, but instead can start learning continually from the very first task.
In our approach, we save a small learned task embedding vector for each task, while in LOTUS~\cite{wan2024lotus}, skills are represented by separate networks. 
}

\notnew{Similar to our current work, hypernetworks are utilized by some works in robotics CL.}
In ~\cite{huang2021continual}, the dynamics model of a manipulator is generated by a hypernetwork, and multiple tasks are learned continually with model-based RL.
In a similar vein, a hypernetwork-generated agent learns multiple manipulation tasks with model-free RL~\cite{schulman2017proximal} in~\cite{schopf2022hypernetwork}. 
An approach to supervised continual LfD is proposed in~\cite{auddy2022continual}, where the parameters of a NODE~\cite{chen2018neural} are generated with a hypernetwork that learns a sequence of LfD tasks.
\notnew{
Our present work also uses hypernetworks, but instead of RL~\cite{huang2021continual, schopf2022hypernetwork}, we rely on \emph{supervised} continual LfD. We evaluate on real-world tasks and assess CL performance on longer task sequences ranging from 7-26 tasks compared to 5-6 tasks in ~\cite{huang2021continual, schopf2022hypernetwork}.
}
\notnew{
In contrast to the continual LfD approach of~\cite{auddy2022continual}, our hypernetwork generates two neural networks of the \snode{} (instead of a single NODE network). This leads to stable motion and improved CL performance. Additionally, we propose a regularization technique to improve hypernetwork training efficiency. We evaluate scalability on high-dimensional tasks (up to 32D), and a longer sequence of 9 real-world LfD tasks \emph{vis-\`a-vis} 4 tasks in the previous work.
}
As far as we are aware, 
ours is the first work that highlights the positive influence of motion stability on CL performance.

\section{Background}
\label{sec:back}

In this section, we briefly describe the fundamentals of our proposed system.
Our new contributions are described later in \mysection{sec:methods}.
As we focus on neural network-based CL and LfD, the discussion here also covers only such techniques.

\subsection{Training a NODE}\label{sec:node_train}

A Neural ODE solver (NODE)~\cite{chen2018neural} is a neural network that learns a dynamical system from observations. Here, a neural network $\hat{\mathbf{f}}_\uptheta(\mathbf{x}): \mathbb{R}^n \rightarrow \mathbb{R}^n$ with parameters $\uptheta$ represents a dynamical system. By integrating this function, an approximate solution to the ODE system is obtained~\cite{auddy2022continual} as

\begin{equation}
    \hat{\mathbf{x}}_t = \newsecond{\mathbf{x}_0} +  \int_{0}^{t} \hat{\mathbf{f}}_\uptheta(\hat{\mathbf{x}}_\tau) \mathrm{d}\tau
    \label{eq:node_int}
\end{equation}
\noindent where $\newsecond{\mathbf{x}_0}$ is the initial state. 
NODE is trained using a dataset $\mathcal{D}$ of $N$ demonstration trajectories 
$\newsecond{\{\mathbf{x}_{1:T}^n\}_{n=1}^{N}}$
where each trajectory $\mathbf{x}^{n}_{1:T}$ is a sequence of $T$ points $\mathbf{x}^{n}_t \in \mathbb{R}^d$. 
In each training iteration, a short contiguous segment named $\mathcal{D}_\sigma$ of length $T_\sigma$ (where $T_\sigma \ll T$) is randomly drawn from the $N$ trajectories in $\mathcal{D}$. 
Thus, $\mathcal{D}_\sigma$ is a tiny subset of $\mathcal{D}$, containing $N$ trajectories each of length $T_\sigma$, such that the starting location of $\mathcal{D}_\sigma$ is at a random temporal location within $\mathcal{D}$. 
Given an input $\mathbf{x}_t^n$, NODE predicts the derivatives of the input that are then numerically integrated to produce a trajectory $\hat{\mathbf{x}}_{1:T_\sigma}^{n}$.
The parameters $\uptheta$ of the NODE can then be learned by minimizing the following loss \cite{auddy2022continual}:
\begin{equation}
    \mathcal{L} = \frac{1}{2} \sum_{n=1}^{N}\sum_{t=1}^{T_\sigma} \Vert \mathbf{x}^{n}_t - \hat{\mathbf{x}}^{n}_t\Vert^2_2
    \label{eq:node_mse_loss}
\end{equation}
NODE uses the computationally-efficient \emph{adjoint method}~\cite{pontryagin2018mathematical,chen2018neural} to compute gradients. Note that the training data only consists of states $\mathbf{x}_t^n$, and ground-truth derivatives of these states are not required.

\myequation{eq:node_mse_loss} cannot be directly used to learn orientations as they do not reside in Euclidean space. Hence we follow~\cite{ude2014orientation, huang2020toward, auddy2022continual}, and first project unit orientation quaternions $\newsecond{\mathbf{q}_t \in \mathbb{S}^3}$ into a locally-Euclidean tangent space with the \emph{logarithmic map}~\cite{saveriano2019merging}.
The rotation vectors $\newsecond{\mathbf{r}_t \in \mathbb{R}^3}$ produced by this operation are learned with~\myequation{eq:node_mse_loss}.
The predicted rotation vectors are then projected back to corresponding quaternions on the hypersphere using the \emph{exponential map}~\cite{saveriano2019merging}.

\subsection{Stability via a jointly learned Lyapunov function}

A basic NODE trained using \myequation{eq:node_mse_loss} does not guarantee stability of the predicted motion. If learning is not successful or the initial state is too different from the demonstrations, then it is possible for the predicted trajectory to diverge from the goal. 

To solve this problem, the authors of~\cite{kolter2019learning} propose to jointly learn a dynamics model and a \emph{Lyapunov} function that ensures %
exponential
stability. 
In addition to the nominal dynamics model 
$\hat{\mathbf{f}}_{\uptheta}(\mathbf{x}): \mathbb{R}^n \rightarrow \mathbb{R}^n$, 
let $V_\upgamma(\mathbf{x}): \mathbb{R}^n \rightarrow \mathbb{R}$ (parameterized by $\upgamma$) denote a Lyapunov function, which is a positive definite function, such that $V_\upgamma(\mathbf{x}) \ge 0$ for $\mathbf{x} \ne 0$ and $V_\upgamma(0) = 0$. The projection of $\hat{\mathbf{f}}_{\uptheta}$ that satisfies the condition
\begin{equation}
    \nabla V_\upgamma(\mathbf{x})^\newsecond{\top} \hat{\mathbf{f}}_{\uptheta}(\mathbf{x}) \le -\alpha V_\upgamma(\mathbf{x})
    \label{eq:stability_condition}
\end{equation}
\noindent ensures a stable dynamics system, where $\alpha \geq 0$ is a constant.
The following function is guaranteed to produce stable trajectories 
 \cite{kolter2019learning}:
\begin{align}
    &\mathbf{f}_{\upphi}(\mathbf{x}) = \mathrm{Proj}\left( \hat{\mathbf{f}}_{\uptheta}(\mathbf{x}), \{f: \nabla V_\upgamma(\mathbf{x})^\newsecond{\top} f(\mathbf{x}) \le -\alpha V_\upgamma(\mathbf{x})\}\right) \nonumber \\ 
         &= \hat{\mathbf{f}}_{\uptheta}(\mathbf{x}) - \nabla V_\upgamma(\mathbf{x}) \frac{\mathrm{ReLU}(\nabla V_\upgamma(\mathbf{x})^\newsecond{\top} \hat{\mathbf{f}}_{\uptheta}(\mathbf{x})) + \alpha V_\upgamma(\mathbf{x})}{||\nabla V_\upgamma(\mathbf{x})||^2_2}
    \label{eq:snode}
\end{align}
where $\upphi=\{\uptheta,\upgamma\}$ represents the parameters of the nominal dynamics model and the Lyapunov function taken together.
The Lyapunov function $V_\upgamma$ is modeled with an input-convex neural network (ICNN)~\cite{amos2017input}.
During training, both the parameters of the nominal dynamics model $\hat{\mathbf{f}}_\uptheta(\mathbf{x})$ and the Lyapunov function $V_\upgamma$ are learned jointly \cite{kolter2019learning} with \myequation{eq:node_mse_loss}.
We refer to the stable dynamics model represented by \myequation{eq:snode} as \emph{stable} NODE (\snode{}).

Stability enforced with a separate loss~\cite{chow2018lyapunov, richards2018lyapunov} or trained \emph{a posteriori} may be difficult to verify~\cite{kolter2019learning, lawrence2020almost}. The widely-adopted technique of jointly training the nominal dynamics model and the Lyapunov function~\cite{kolter2019learning, lawrence2020almost,sochopoulos2024learning} is verifiably stable and results in the trainable parameters reaching local minima where both accuracy and stability are achieved.
Note that though the nominal dynamics model $\hat{\mathbf{f}}_{\uptheta}(\cdot)$ and the Lyapunov function $V_\upgamma(\cdot)$ are separate functions with their own inputs and outputs, the entire system can be trained together by using \myequation{eq:snode} to compute the loss in \myequation{eq:node_mse_loss}. Ground-truth for the outputs of $\hat{\mathbf{f}}_{\uptheta}(\cdot)$ and $V_\upgamma(\cdot)$ are not required.

\subsection{Hypernetworks}\label{sec:back_hn}

A hypernetwork is a neural network that generates another neural network~\cite{ha2017hypernetworks}. 
The generated network (\emph{task-learner}) performs the actual task under consideration.
The hypernetwork's input is a vector that is also trainable.
After being generated, the task-learner is fed with the input data, the training loss is computed using its output, and gradients are backpropagated through the entire system. 
Note that the task-learner parameters are simply outputs of the hypernetwork and are not trainable.

In CL~\cite{von2019continual}, hypernetwork parameters are protected from catastrophic forgetting~\cite{parisi2019CL_survey} with regularization, and the hypernetwork input vectors are called \emph{task embeddings}~\cite{von2019continual}.
While learning the $m^\upth$ task in a sequence of CL tasks,
consider that a hypernetwork with parameters $\mathbf{h}$ is fed with a newly initialized task-embedding $\mathbf{e}^m$, and generates the parameters $\mathbf{\uptheta}^m$ of a task learner.
To optimize $\{\mathbf{h}, \mathbf{e}^m\}$, a two-stage process is followed \cite{von2019continual, auddy2022continual}. First, a candidate change $\Delta\mathbf{h}$ is determined such that the task-specific loss $\mathcal{L}^m$ for the $m^\upth$ task is minimized. $\mathcal{L}^m$ depends on the task-learner parameters $\uptheta^m$ (generated by the hypernetwork) and $\mathbf{x}^m$, the data for the $m^\upth$ task. Thus,
\begin{equation}
	\mathcal{L}^m = \mathcal{L}^m(\uptheta^m = \mathbf{f}(\mathbf{e}^m, \mathbf{h}), \mathbf{x}^m)
	\label{eq:hn_loss_step1}
\end{equation}
\noindent where $\mathbf{f}$ is the hypernetwork function.
Next, the regularized loss $\tilde{\mathcal{L}}^m$ is minimized to optimize \{$\mathbf{h}, \mathbf{e}^m$\}~\cite{von2019continual, auddy2022continual}:
\begin{align}
	\tilde{\mathcal{L}}^m = &\mathcal{L}^m(\uptheta^m = \mathbf{f}(\mathbf{e}^m, \mathbf{h}), \mathbf{x}^m) \nonumber \\
	& + \cfrac{\beta}{m-1} \sum\limits^{m-1}_{l=0}\left\vert\left\vert\mathbf{f}(\mathbf{e}^l, \mathbf{h}^*) - \mathbf{f}(\mathbf{e}^l, \mathbf{h}+\Delta\mathbf{h})\right\vert\right\vert^2
	\label{eq:hn_loss_step2}
\end{align}
\noindent Here, $\beta$ is a constant and $\mathbf{h}^*$ denotes the hypernetwork parameters before learning the $m^\upth$ task. For each task, a new task embedding vector is learned and then frozen for regularizing the learning of future tasks.

As the task learner parameters $\uptheta^m$ are outputs of the final layer of the hypernetwork, this last layer can become quite large if $\uptheta^m$ is large. 
To keep the hypernetwork size small, \emph{chunked} hypernetworks have been proposed~\cite{von2019continual},
where $\uptheta^m$ is produced in segments called \emph{chunks}. %
In addition to the task embedding vector $\mathbf{e}^m$, chunked hypernetworks use a set of additional inputs called \emph{chunk embedding vectors}, which are also trainable. A separate task embedding vector is learned for each task, while a single set of chunk embedding vectors is shared across all tasks and is regularized in the same way as the hypernetwork parameters  
(see~\cite{von2019continual} for further details).

The advantage of chunked hypernetworks is that the final hypernetwork layer can be reasonably small, resulting in an overall smaller network size. Based on the dimensions of the input vectors and layer sizes, the size of a chunked hypernetwork can be comparable to the network it generates or even smaller~\cite{auddy2022continual}. A smaller network may also be easier to train. 
However, a small size can also be a drawback and result in a less expressive 
chunked hypernetwork that remembers fewer tasks than a regular hypernetwork~\cite{auddy2022continual}.
Overall, hypernetworks have exhibited good CL performance in diverse scenarios~\cite{von2019continual,huang2021continual,auddy2022continual,schopf2022hypernetwork,ehret2020continual}. A single hypernetwork can learn multiple tasks, does not store or replay training data from past tasks, and only grows minimally with new tasks~(due to storage of the small task embeddings).

\begin{figure*}[t!]
  \centering
  \subfloat[
  \newer{
    Our proposed clock-augmented \snode{} takes an additional clock input $c_t$ that is appended to $\mathbf{x}_t$ to create an augmented state $\hat{\mathbf{x}}_t$. The clock input $c_t\in{[0,1]}$ is independent of $\mathbf{x}_t$ and evolves linearly. $\hat{\mathbf{x}}_t$ is an input to the nominal dynamics model $\hat{\mathbf{f}}_\uptheta$ and the parameterized Lyapunov function $V_\gamma$. This improves the accuracy of predictions without affecting stability.
    }
  ]{\includegraphics[width=0.48\columnwidth]{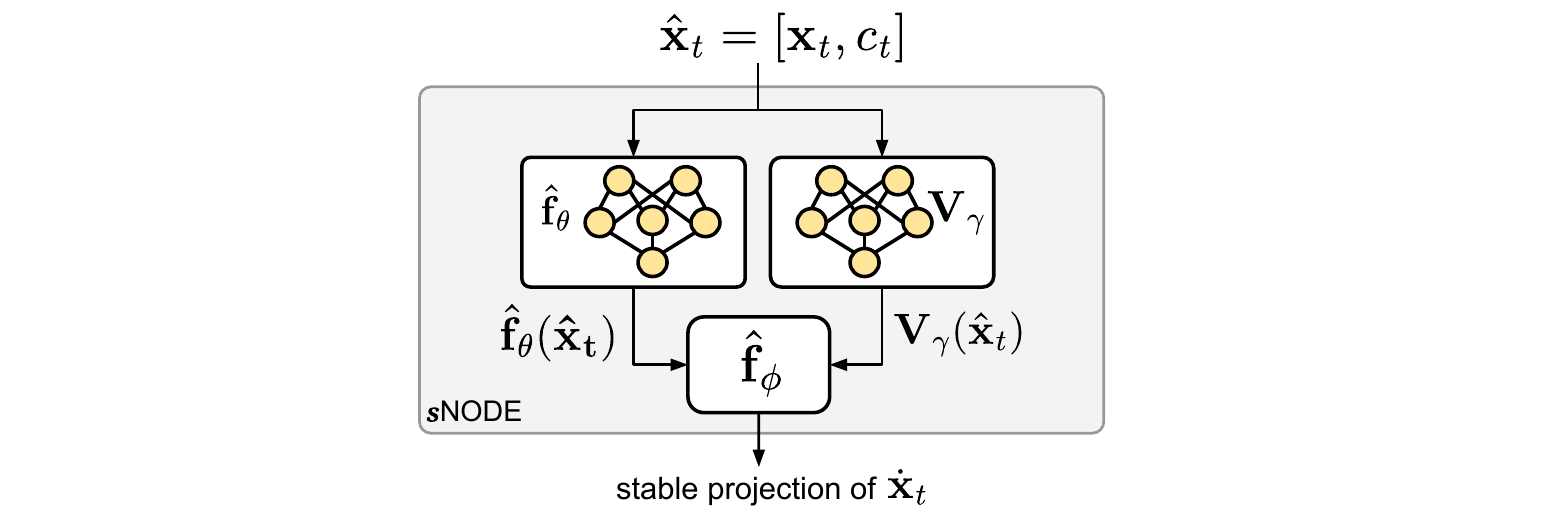}
  \label{fig:snode_a}}
  \hspace{5pt}
  \subfloat[
    \newer{
    On real-world tasks involving 6-DoF robot poses, \snode{} can learn position and orientation simultaneously by projecting the orientation quaternions into rotation vectors with the \texttt{Log} map, learning positions and rotation vectors together in Euclidean space, and then projecting the predicted rotation vectors back into quaternions with the \texttt{Exp} map.
    }
  ]{\includegraphics[width=0.48\columnwidth]{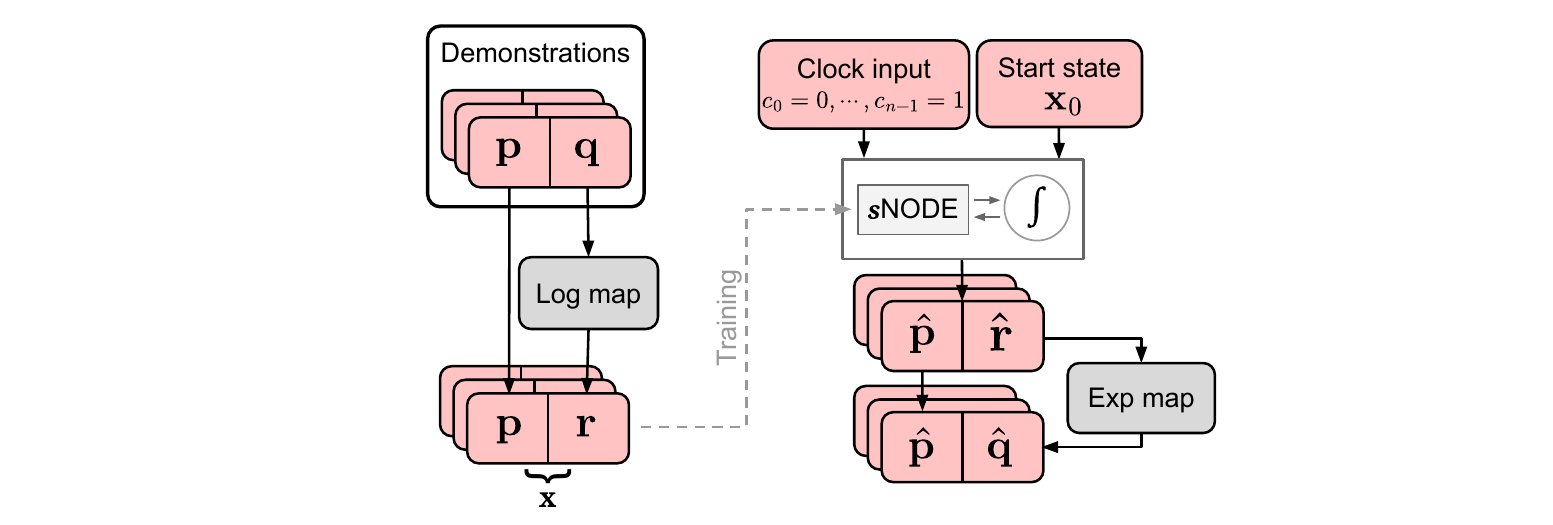}
  \label{fig:snode_b}}
  \caption[]{ 
    Our proposed clock-augmented stable NODE (\snode{}) and its usage to jointly learn position and orientation for real-world tasks.
  }
  \label{fig:snode}    
\end{figure*}

\section{Methods}
\label{sec:methods}

\newsecond{
We augment the \snode{} with a clock input to improve accuracy. Our primary contributions include continual learning hypernetworks that generate these clock-augmented \snode{}s, and a stochastic hypernetwork regularization technique. 
Together, these contributions yield accurate and stable trajectories, improved CL performance, and improved training efficiency.
}%

\subsection{Stable NODE with clock input}
\label{sec:method_snode_c}

We introduce an additional input to the \snode{} model of~\cite{kolter2019learning} that results in more accurate predictions.
The new input $c_t$ is referred to as a \emph{clock}, and, as illustrated in \myfigure{fig:snode_a}, it is provided to both the nominal dynamics model $\hat{\mathbf{f}}_\uptheta$ and the Lyapunov function $V_\upgamma$. 
The clock signal $c_t$ \textcolor{black}{is an extra state of the system, designed to be bounded and to evolve linearly from 0 to 1 in $T_{\sigma}$ steps. This is obtained by integrating}
\textcolor{black}{
  \begin{equation}
    \dot{c}_t = \begin{cases}
         k = \dfrac{1}{T_{\sigma}}, &  c_t \leq 1,\\
         0, & \text{otherwise}
    \end{cases},
    \label{eq:clock_signal}
  \end{equation}
where $c_0 = 0$ and $T_{\sigma}$ is the length of the motion. The clock signal being independent of $\mathbf{x}_t$, it is possible} to impose stability using a condition similar to~\myequation{eq:stability_condition}\textcolor{black}{, by considering the augmented state $\hat{\mathbf{x}}_t = [\mathbf{x}_t, c_t]$, where $x_t$ is the end-effector position and orientation and $c_t$ is the clock signal}. %
  
The input layer of the neural network representing the nominal dynamics model $\hat{\mathbf{f}}_\uptheta$ is modified to accept the augmented input \textcolor{black}{(state)} $\hat{\mathbf{x}}_t$. %
We leave the output layer of $\hat{\mathbf{f}}_\uptheta$ unchanged and 
simply append a constant $k$ to the output (as $c_t$ evolves linearly, $\dot{c}_t = k$ is a constant \textcolor{black}{defined in \myequation{eq:clock_signal}}):
\begin{equation}
    \hat{\mathbf{f}}_\uptheta(\hat{\mathbf{x}}_t) =  \left[\dot{x}_{0_t}, \dot{x}_{1_t}, \cdots, \dot{x}_{{n-1}_t}, \dot{c}_t=k \right]^\top %
\end{equation}
This change 
enables the combination of $\hat{\mathbf{f}}_\uptheta(\hat{\mathbf{x}}_t)$ with the gradient of the Lyapunov function (in \myequation{eq:snode}), which is now defined as
\begin{equation}
    \nabla V(\hat{\mathbf{x}}_t) = \left[ \frac{\partial V}{\partial x_{0_t}}, \frac{\partial V}{\partial x_{1_t}}, \cdots \frac{\partial V}{\partial x_{{n-1}_t}}, \frac{\partial V}{\partial{c}_t} \right]^\top
\end{equation}
Since the Lyapunov function $V_\upgamma$ produces a scalar output, we only modify the input layer of the ICNN (that models the Lyapunov function) to accept an additional input 
and leave the output unchanged. 
\newsecond{%
We present quantitative results in Figs. \ref{fig:clock_noclock}(a) and \ref{fig:iflow_node_snode}, and qualitative results in \myfigure{fig:clock_noclock}(b) showing that the input disambiguation performed by the additional clock input improves prediction accuracy. 
The clock input's positive effect on accuracy is discussed in detail in Sec. VII of the supplementary materials.
}
For notational convenience, in the remainder of this paper, we refer to our clock-augmented stable node simply as \snode{}, and use it in all successive experiments. Where necessary we disambiguate between \snode{} models with and without the clock input.

\begin{figure}[t!]
  \centering
  \subfloat[\newer{\hnsnode{}: Parameters $\uptheta$, $\gamma$ of the nominal dynamics model $\hat{\mathbf{f}}_{\uptheta}$ and Lyapunov function $V_\gamma$ of the \snode{} are generated by the last hypernetwork layer. A trainable task embedding (TE) is learned for each task and frozen.}]{\includegraphics[width=\columnwidth]{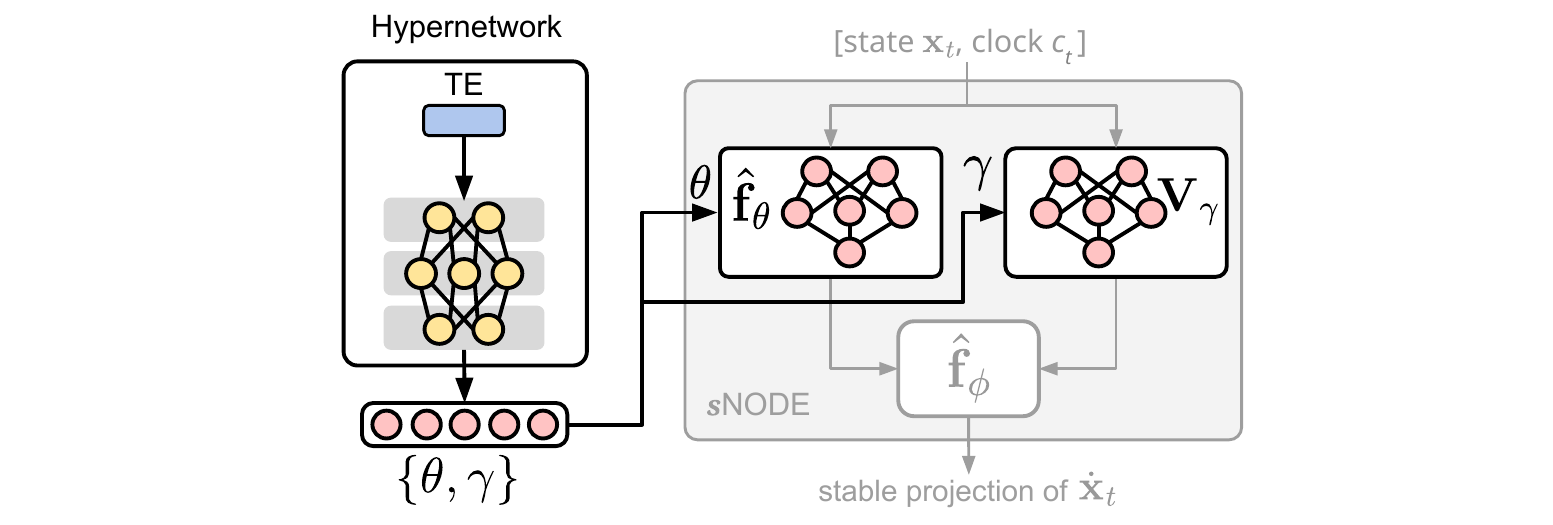}
  \label{fig:hn_chn_a}}
  \\
  \subfloat[\newer{\chnsnode{}: Trainable chunk embeddings (CE) help to generate parameters $\uptheta$, $\gamma$ of the \snode{} in \emph{chunks}, allowing for a smaller hypernetwork. CE is shared between tasks and a trainable task embedding vector (TE) is learned for each task and frozen.}]{\includegraphics[width=\columnwidth]{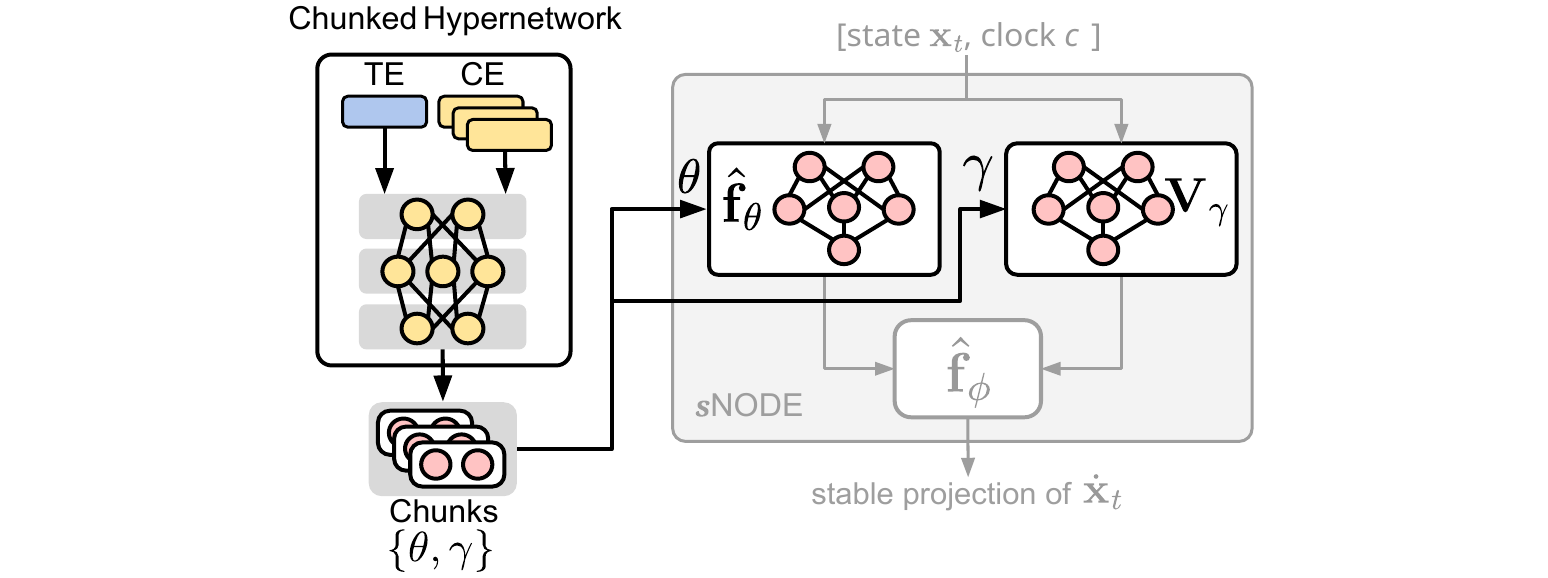}
  \label{fig:hn_chn_b}}
  \caption[]{ 
    Our proposed hypernetworks for stable continual LfD.
    Task-specific parameters are shown by~\legendsquare{myblue}, regularized (task-independent) parameters by~\legendsquare{myyellow}, and non-trainable outputs by~\legendsquare{mypink}. 
    The \snode{} architecture for (a), (b) is the same as in \myfigure{fig:snode_a}.
    Note that contrary to a stand-alone \snode{}, parameters of the \snode{} here are outputs of the hypernetwork and not trainable directly.
  }
  \label{fig:hn_chn}    
\end{figure}

\medskip
For real-world LfD tasks, we learn position and orientation simultaneously. 
As shown in \myfigure{fig:snode_b}, human demonstrations (training data) consist of position trajectories ~$\mathbf{p} = \{\mathbf{p}_0, \cdots, \mathbf{p}_t, \cdots, \mathbf{p}_{T-1} \}$ and orientation trajectories in the form of unit quaternion sequences \newsecond{\mbox{$\mathbf{q} = \{\mathbf{q}_0, \cdots, \mathbf{q}_t, \cdots, \mathbf{q}_{T-1} \}$}}, where each $\mathbf{p}_t \in \mathbb{R}^3$ and $\mathbf{q}_t \in \mathbb{S}^3$. Quaternion trajectories are converted to trajectories of rotation vectors $\mathbf{r} = \{\mathbf{r}_0, \cdots, \mathbf{r}_t, \cdots, \mathbf{r}_{T-1} \}$, where $\mathbf{r}_t \in \mathbb{R}^3$, with the \emph{logarithmic map} $\mathrm{Log}(\cdot)$, as discussed in \mysection{sec:node_train}.
As rotation vectors lie in the local Euclidean tangent space they can be combined with Euclidean positions to form trajectories of 6D vectors $\mathbf{x} = \{ \mathbf{x}_0, \cdots, \mathbf{x}_t, \cdots, \mathbf{x}_{T-1} \}$, where $\mathbf{x}_t = (\mathbf{p}_t, \mathbf{r}_t)$. These are used to train the \snode{} using \myequation{eq:node_mse_loss}.
We found it beneficial to scale the rotation vectors by a constant before concatenating them with positions.
During inference, the orientation component is converted to the quaternion form with the \emph{exponential map} $\mathrm{Exp}(\cdot)$ (see \mysection{sec:node_train}).
Simultaneously learning position and orientation with a single model saves parameter storage and training effort while exploiting potential synergies between the trajectories. In preliminary tests, we found that learning position and orientation with a single model reduces prediction errors for both position and orientation compared to using two separate models.

\subsection{Hypernetworks that generate stable NODE}
We propose regular and chunked hypernetworks that generate the parameters of \emph{two} neural networks that constitute our clock-augmented \snode{} (see \mysection{sec:method_snode_c}): 
\begin{enumerate*}[label=(\roman*)]
	\item a nominal dynamics model, and
	\item a parameterized Lyapunov function~\cite{kolter2019learning}.
\end{enumerate*}
In experiments, the overall size of our hypernetwork$\rightarrow$\snode{} models is kept roughly the same as the hypernetwork$\rightarrow$NODE models of~\cite{auddy2022continual} for a fair comparison.
\smallskip

\noindent\textbf{Hypernetwork$\rightarrow$\snode{} (HN)}:
 As shown in \myfigure{fig:hn_chn_a}, a regular hypernetwork generates the parameters of a clock-augmented \snode{} $\hat{\mathbf{f}}_\upphi$, (see \myequation{eq:snode} and \mysection{sec:method_snode_c}), where $\upphi=\{\uptheta, \gamma \}$. 
 Here, $\uptheta$ represents the parameters of the nominal dynamics model $\hat{\mathbf{f}}_\uptheta$, and $\gamma$ represents the parameters of the Lyapunov function $V_\upgamma$. The parameters $\upphi$ are generated in their entirety from the final layer of the hypernetwork. 
 In the two-step hypernetwork optimization process (Eqs.~\ref{eq:hn_loss_step1} and \ref{eq:hn_loss_step2}), the task-specific loss $\mathcal{L}^m$ for task $m$ is represented by \myequation{eq:node_mse_loss}, and is computed using only the demonstrations provided for this task.
In the remainder of this paper, we refer to this model, where the \snode{} is generated by a regular hypernetwork, as \hnsnode{}.
In our experiments, we compare against the regular hypernetwork model proposed by \cite{auddy2022continual}, and refer to it as \hnnode{}.
\smallskip

\noindent\textbf{Chunked Hypernetwork$\rightarrow$\snode{} (CHN)}:
As shown in \myfigure{fig:hn_chn_b}, a chunked hypernetwork generates the parameters $\upphi=\{\uptheta, \gamma \}$ of a clock-augmented \snode{} $\hat{\mathbf{f}}_\upphi$ (see \myequation{eq:snode} and \mysection{sec:method_snode_c}) in segments known as \emph{chunks}. We assemble the nominal dynamics model $\hat{\mathbf{f}}_\uptheta$ and the Lyapunov function $V_\upgamma$ from the generated chunks. 
The two-step optimization process (Eqs.~\ref{eq:hn_loss_step1} and \ref{eq:hn_loss_step2}) for a chunked hypernetwork also uses \myequation{eq:node_mse_loss} as the task-specific loss $\mathcal{L}^m$ for task $m$.
We refer to this model as \chnsnode{} and treat the \chnnode{} model from \cite{auddy2022continual} as a comparative baseline.
The advantages of chunked hypernetworks \emph{vis-\`a-vis} regular hypernetworks, discussed in~\mysection{sec:back_hn}, remain relevant for \chnsnode{}.

\subsection{Stochastic regularization in hypernetworks}
\label{sec:method_hn_sampled_reg}

Hypernetworks are a suitable choice for CL~\cite{von2019continual, huang2021continual, auddy2022continual, brahma2021hypernetworks},
but their regularization cost increases for each new task, leading to a cumulative training cost of $\mathcal{O}(N^2)$ for $N$ tasks.
This occurs due to iterating over stored task embeddings of all previous tasks~(\myequation{eq:hn_loss_step2}).
A solution to this problem, proposed by~\cite{von2019continual}, is to use a random subset (of fixed size) of past task embeddings for regularization. Thus, for learning the $m^\upth$ task, the hypernetwork loss function in the second optimization step (\myequation{eq:hn_loss_step2}) becomes
\begin{align}
	\tilde{\mathcal{L}}^m = & \mathcal{L}^m(\uptheta^m, \mathbf{x}^m) \nonumber \\
	& + \cfrac{\beta}{|\mathcal{K}|} \sum\limits^{\mathcal{K} \sim  \mathrm{U}(\mathcal{E}_{m-1})}_{\mathbf{e}^l \subset \mathcal{K}}\left\vert\left\vert\mathbf{f}(\mathbf{e}^l, \mathbf{h}^*) - \mathbf{f}(\mathbf{e}^l, \mathbf{h}+\Delta\mathbf{h})\right\vert\right\vert^2
	\label{eq:hn_loss_step2_fix1}
\end{align}
\noindent where $\mathcal{K}$ denotes the random subset of task embeddings sampled uniformly from all past task embeddings $\mathcal{E}_{m-1} = \{\mathbf{e}^0, \cdots, \mathbf{e}^{m-1}\}$. Other symbols have the same meaning as in \myequation{eq:hn_loss_step2}. $\vert\mathcal{K}\vert$ is fixed, and as long as $|\mathcal{K}| <= m-1$ (i.e. until more tasks than the size of $\mathcal{K}$ have been learned), $\mathcal{K}$ simply includes all past task embeddings. 
This helps to set an upper bound on the time and effort for hypernetwork training: the cumulative training time increases quadratically till $|\mathcal{K}| <= m-1$, after which it becomes linear.

To further reduce the computational overhead, we propose to remove the summation operation in \myequation{eq:hn_loss_step2_fix1}. In each training iteration, we uniformly sample a single task embedding from the set of all past task embeddings (i.e. $|\mathcal{K}|=1$) and use it for regularization:
\begin{align}
	&\tilde{\mathcal{L}}^m = \mathcal{L}^m(\uptheta^m, \mathbf{x}^m) + \beta\left\vert\left\vert\mathbf{f}(\mathbf{e}^k, \mathbf{h}^*) - \mathbf{f}(\mathbf{e}^k, \mathbf{h}+\Delta\mathbf{h})\right\vert\right\vert^2
	\label{eq:hn_loss_step2_fix2} \\
	& \text{where }\mathbf{e}^k \sim \mathrm{U}(\mathcal{E}_{m-1}) \nonumber
\end{align}
With this change, the cumulative training cost for $N$ tasks becomes $\mathcal{O}(N)$ instead of $\mathcal{O}(N^2)$. This cumulative cost is achieved from the very first task, unlike \cite{von2019continual}.
We empirically show the benefits of this approach and also discuss its limitations later in \mysection{sec:results_stoch_reg}.

Note that the hypernetwork training procedure with our proposed approach in \myequation{eq:hn_loss_step2_fix2} is identical to that of Eqs.~\ref{eq:hn_loss_step2} and \ref{eq:hn_loss_step2_fix1}. In all cases, only the training data of the newest task is used without storing or replaying past data, and regularization occurs in the space of hypernetwork outputs. 
The difference is that we do not iterate over saved task embeddings and 
thereby keep the regularization effort for each task nearly the same.

\section{Experimental Setup}
\label{sec:experiment_setup}

\subsection{Datasets}
\label{sec:experiments_datasets}

\noindent\textbf{Two-dimensional LfD tasks}: 
We use two different LfD datasets containing two-dimensional trajectories.
Similar to prior works~\cite{urain2020imitationflow,saveriano2020energy,ravichandar2020recent,blocher2017learning,auddy2022continual}, we use the popular \emph{LASA}~\cite{khansari2011learning} LfD benchmark (\lasa{2}) in our evaluations. We use the same 26 tasks as~\cite{auddy2022continual} and choose this dataset because its large number of diverse LfD tasks is particularly suitable for evaluating CL performance when all tasks are learned sequentially. Refer to~\cite{khansari2011learning, auddy2022continual} for further details.

Additionally, we use \emph{HelloWorld}~\cite{auddy2022continual}~(\hw{}), an LfD dataset containing kinesthetic demonstrations of writing the 7 unique lowercase letters in ``\emph{hello world}'', collected with a robot.
\hw{} features more complex trajectories than \lasa{2}, including self-loops and instances where the initial and goal positions are close, offering scenarios useful for evaluating the stability-accuracy balance of stable dynamical models (see \mysection{sec:results_snode}). Further details of this dataset can be found in~\cite{auddy2022continual}.
\smallskip

\noindent\textbf{High-dimensional LfD tasks}:
We propose a suite of high-dimensional LfD datasets based on \lasa{2} to address the need of benchmarks for assessing scalability of continual LfD to high-dimensional trajectories.
We create three datasets \lasa{8}, \lasa{16}, and \lasa{32} of 8-, 16- and 32-dimensional trajectories respectively, by concatenating multiple unique tasks chosen uniformly from \lasa{2} into a single LfD task. For example, a single task in \lasa{32} is created by concatenating 16 tasks from \lasa{2}. 
Each high-dimensional dataset is a sequence of 10 tasks, where each task contains 7 demonstrations, and each demonstration is a sequence of 1000 points. The dimension of the points are 8, 16, and 32 for \lasa{8}, \lasa{16}, and \lasa{32} respectively. 
\smallskip

\noindent\textbf{Real-world LfD tasks}:
For real-world evaluation, we consider LfD tasks where both the position and orientation of the robot's end effector vary. 
We expand the \emph{RoboTasks} dataset~\cite{auddy2022continual} from 4 to 9 real-world LfD tasks, creating \emph{RoboTasks9} (\robot{}).
The tasks in this dataset are: 
\begin{enumerate*}[label=(\roman*)]
    \item \emph{box~opening}: lid of a box is opened,
    \item \emph{bottle~shelving}: a bottle is placed on a shelf,
    \item \emph{plate~stacking}: a plate is placed on a table,
    \item \emph{pouring}: coffee beans are poured into a container,
    \item \emph{mat~folding}: a mat is folded in half,
    \item \emph{navigating}: an object is carried between obstacles,
    \item \emph{pan~on~stove}: a pan is transferred from a hanging position to a table,
    \item \emph{scooping}: coffee beans are scooped with a spatula, and
    \item \emph{glass~upright}: a wine glass on its side is set upright.
\end{enumerate*}
Tasks (v)-(ix) are created by us in this paper (all tasks are illustrated in \myfigure{fig:intro}b).
Each task contains 9 demonstrations provided kinsethetically by a human. Each demonstration is a sequence of 1000 steps, and in each step, the robot's position $\mathbf{p} \in \mathbb{R}^3$ and orientation $\mathbf{q} \in \mathbb{S}^3$ are recorded.
\robot{} allows us to evaluate our approach in real-world scenarios, and the sequence of 9 tasks is more suitable for assessing CL performance than the shorter sequences of 4--6 tasks in prior work~\cite{huang2021continual,auddy2022continual,schopf2022hypernetwork}. 

\subsection{Evaluation metrics}
\label{sec:experiments_metrics}

We report metrics for trajectory errors, CL performance and stability.
We measure the difference between predicted trajectories and demonstrations and report the widely-used metrics \emph{Dynamic~Time~Warping~error}~(DTW)~\cite{urain2020imitationflow,jekel2019similarity,auddy2022continual} for position, and \emph{Quaternion~error}~\cite{ude2014orientation,saveriano2019merging,auddy2022continual} for orientation.

\new{
Trajectory errors tell us how accurate predictions are, but they do not capture the different tradeoffs that can be made while learning multiple tasks continually. For example, the naive approach of learning each task with a separate model
results
in low errors and no forgetting, but does not scale well in terms of overall parameter size. 
CL metrics capture multiple performance aspects, offering a comprehensive view of the trade-offs inherent in different approaches.
We follow the existing CL literature~\cite{liu2020generative, wan2024lotus, daruna2021continual, auddy2022continual} and report individual CL metrics and overall CL scores. These metrics are:
}
\begin{enumerate*}[label=(\roman*)]
    \item \emph{ACC}~\cite{diaz2018don}:~average prediction \emph{accuracy};
    \item \emph{REM}~\cite{diaz2018don}:~how well past tasks are \emph{remembered};
    \item \emph{MS}~\cite{diaz2018don}:~growth of \emph{model size} with new tasks;
    \item \emph{SSS}~\cite{diaz2018don}:~growth in \emph{sample storage size} due to retaining training data from past tasks;
    \item \emph{TE}~\cite{auddy2022continual}:~change in \emph{training efficiency} (in terms of time) with new tasks;
    \item \emph{FS}~\cite{auddy2022continual}:~relative \emph{final size} of a model.
\end{enumerate*} 
Each CL metric lies in the range [0,1] with 1 being the best.
The set of individual CL metrics $\mathcal{C}=\{\mathrm{ACC, REM, MS, SSS, TE, FS}\}$, is used~to compute overall metrics $\mathrm{CL_{score}}=\sum_i^{n(\mathcal{C})} c_i$ and $\mathrm{CL_{stability}} = 1 - \sum_i^{n(\mathcal{C})} \mathrm{STDEV}(c_i)$~\cite{diaz2018don}. 
\new{
Details of the CL metrics are provided in~Tab. I in the supplementary materials.}

We also empirically assess stability in two ways. 
\new{First, we initialize a model at a starting point that is different from the demonstration and measure $\Delta_\text{EP}$, the distance between the \emph{end points} of the predicted and demonstrated trajectories. For stable trajectories, $\Delta_\text{EP}$ should be small. 
Second, we roll out trajectories with lengths greater than the demonstration and measure $\Delta_\text{EP+}$, the distance between the final point of the longer predicted trajectory and that of the shorter demonstration. Ideally, a predicted trajectory should always terminate close to the goal irrespective of the rollout length, and $\Delta_\text{EP+}$ should have a low value. Details of our stability metrics and} quantitative results of our empirical stability tests are reported in \mysection{sec:stability}.
\new{
}

\subsection{Baselines}\label{sec:baselines}

We evaluate our approach against CL baselines from all CL families (\emph{dynamic growth}, \emph{replay}, \emph{regularization})~\cite{parisi2019CL_survey}. 
\new{
Note that, corresponding to each CL method, we have 2 baselines: one implemented with NODE~\cite{auddy2022continual}, and another implemented with \snode{}.
}
We compare our proposed hypernetworks (HN/CHN$\rightarrow$\snode{}) against (HN/CHN$\rightarrow$NODE)~\cite{auddy2022continual} and the following CL baselines~\cite{auddy2022continual}:
\begin{enumerate*}[label=(\roman*)]
  \item \textbf{SG}:~each task is learned with a \emph{single} (i.e. dedicated) NODE/\snode{} that is frozen afterwards. This \emph{dynamic growth} setting is an upper performance baseline, as learning a new task does not induce forgetting in previous frozen models;
  \item \textbf{FT}:~a lower performance baseline where a NODE/\snode{} is sequentially \emph{finetuned} on each task;
  \item \textbf{REP}:~training data of each task is stored and \emph{replayed} for training a NODE/\snode{} on each subsequent task via multi-task learning;
  \item \textbf{SI}:~a NODE/\snode{} is sequentially trained on each task and the \emph{Synaptic Intelligence}~\cite{zenke2017SI} regularization algorithm is used to prevent forgetting;
  \item \textbf{MAS}:~another regularization CL approach known as \emph{Memory Aware Synapses~\cite{aljundi2018MAS}} is used to train a NODE/\snode{} sequentially.
\end{enumerate*}
Similar to~\cite{auddy2022continual}, FT, REP, SI, and MAS are task-conditioned by a trainable task-embedding vector, as we need to specify the task to be executed during inference.
\new{
Detailed description of all the baselines is provided in~Tab. II in the supplementary document.
We do not directly compare against approaches such as LOTUS~\cite{wan2024lotus} and CRIL~\cite{gao2021cril} due to differences in data modalities,
but our existing baselines represent the core strategies employed by these methods. For example, the SG baseline learns a separate NODE/\snode{} for each task, and the collection of NODEs/\snode{}s can be considered to be a library of skills similar to that in \cite{wan2024lotus}. Similarly, our REP baseline is similar to the replay-based approach adopted by~\cite{gao2021cril}.
}
Further, as discussed in~\mysection{sec:intro}, we focus on neural network-based \emph{continual} LfD, and therefore do not consider LfD methods not based on neural networks.
Though \snode{} can be substituted by other neural network-based dynamical systems in future work, currently we evaluate only \snode{} and NODE to keep the number of experimental baselines manageable. 
\newcolumntype{C}[1]{>{\centering\arraybackslash\hsize=#1\hsize}X}
\newcolumntype{L}[1]{>{\arraybackslash\hsize=#1\hsize}X}

\section{Results}
\label{sec:experiments_results}

We present the results of our experiments on the datasets described in \mysection{sec:experiments_datasets}. 
Hyperparameters and additional results can be found in Tab. IV. and Sec. VII, respectively, in the supplementary materials.

\subsection{Stable NODE with clock input}
\label{sec:results_snode}

\begin{figure}[t!]
  \centering
  \subfloat[Clock input in \snode{} reduces trajectory errors on LASA~2D and especially on HelloWorld. Overall errors for all tasks are shown (5 independent seeds).]{\includegraphics[width=\columnwidth]{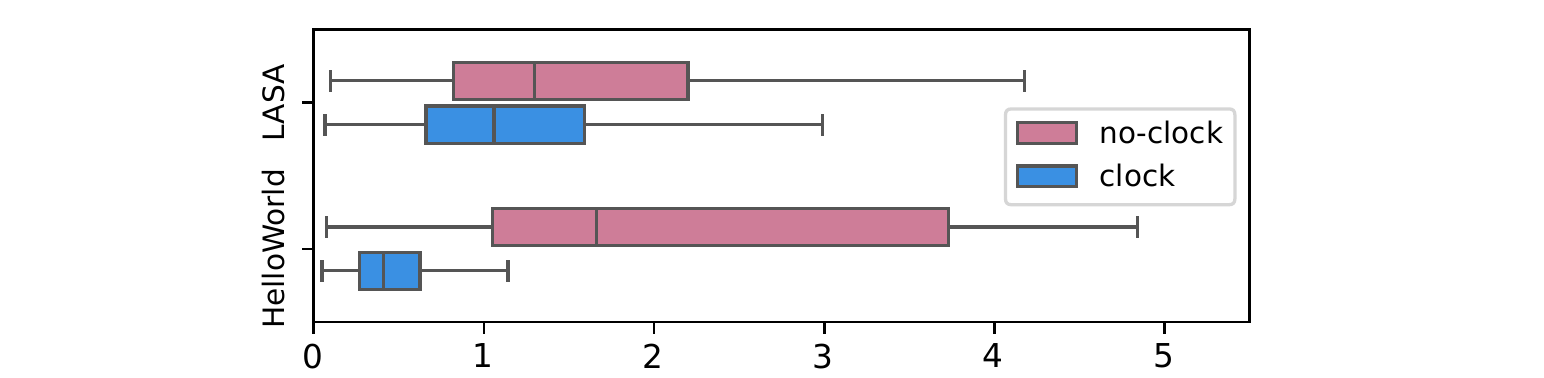}
  \label{fig:clock_noclock_a}} \\
  \subfloat[Qualitative examples on HelloWorld. \snode{} without clock input (\emph{no-clock}) makes mistakes while \snode{} with clock input (\emph{clock}) doesn't. Each row shows a different task, each column shows a different demo of the same task.]{\includegraphics[width=\columnwidth]{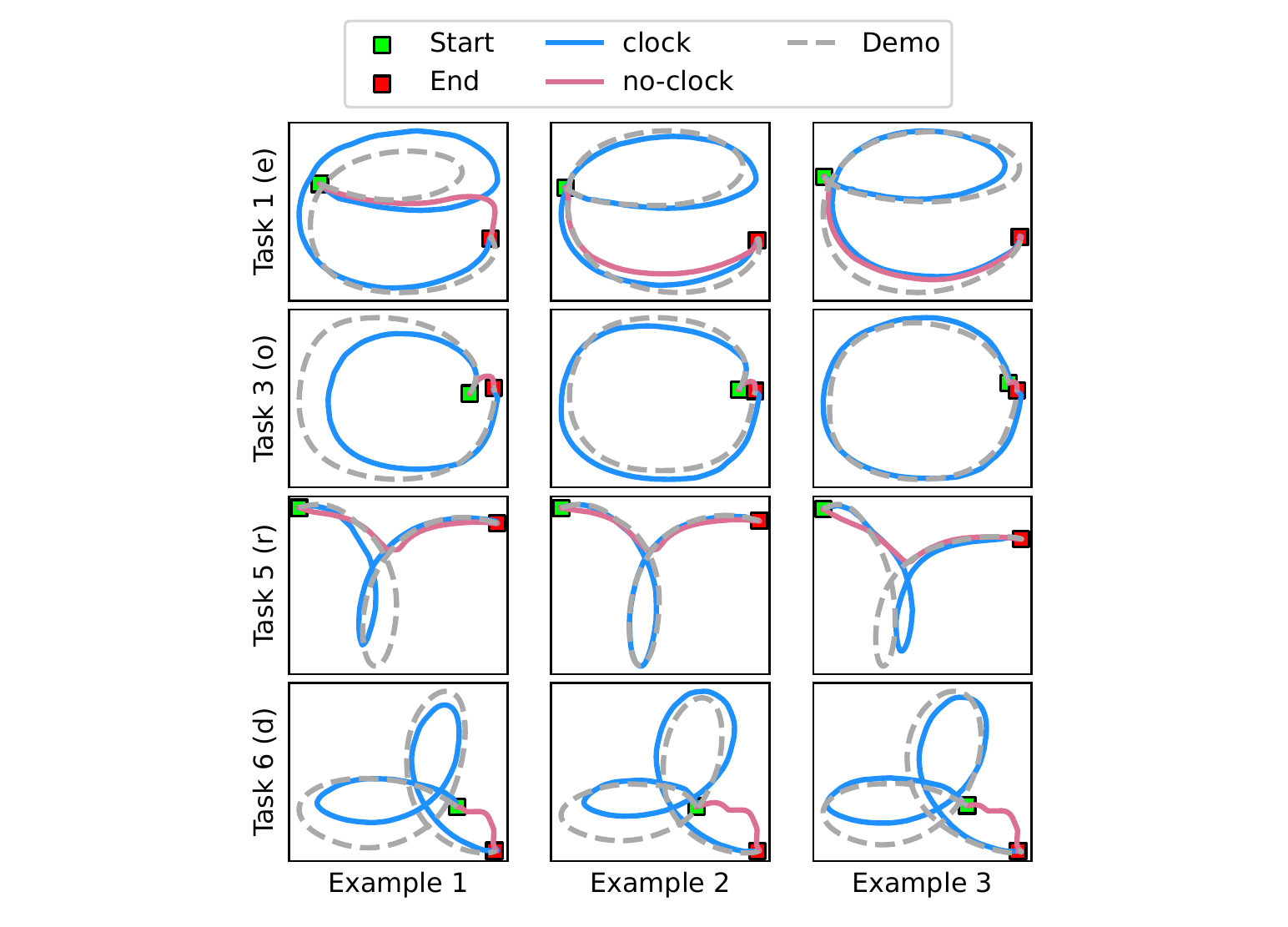}
  \label{fig:clock_noclock_b}}
  \caption[]{ 
    Comparison of \snode{} with and without clock input. 
    }
  \label{fig:clock_noclock}    
\end{figure}

In this non-continual learning experiment, we train a dedicated \snode{} for each task (for both \snode{} with clock and \snode{} without clock) on \lasa{2} and \hw{}, and compare the overall prediction errors.
\myfigure{fig:clock_noclock_a} shows that the errors of our clock-augmented \snode{} are less than \snode{} without clock on both datasets. This difference is more pronounced for \hw{}, which contains more complicated trajectories than \lasa{2}.
The qualitative examples for \hw{} in \myfigure{fig:clock_noclock_b} show that when the initial and goal locations are nearby (letters `o' and `d'), or when the demonstrated trajectory passes close to the goal (letter `e'), the prediction of \snode{} without clock input goes directly to the goal and does not resemble the demonstration. \snode{} without clock input is also unable to account for loops in the demonstration (letters `r' and `d'). 
In all these cases, \snode{} with clock input maintains the desired balance between stability and accuracy, and produces accurate predictions. 
Note that all hyperparameters and stability properties are the same for both kinds of \snode{} in these experiments.
The reported results are obtained with 5 independent seeds.

These examples highlight the benefit of the clock input for \snode{} (see \mysection{sec:method_snode_c}) and similar situations can also manifest in real-world LfD tasks. Due to the superior accuracy of \snode{} with clock, we use it for LfD in all subsequent experiments, and refer to it simply as \snode{} (without mentioning the clock).  
\newer{
In Sec. VII of the supplementary materials, we present an explanation for the improved accuracy due to the clock signal.}

\begin{figure}[t]
    \centering
    \includegraphics[width=0.7\textwidth]{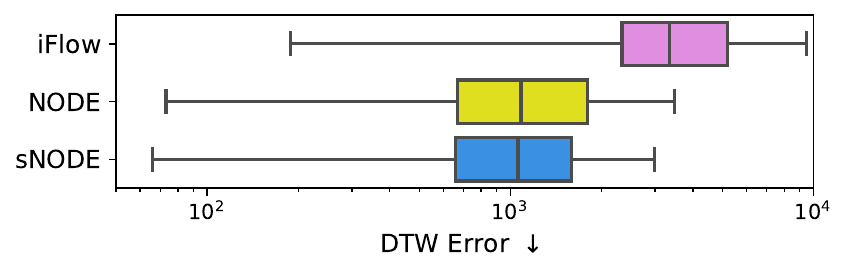}
    \caption{\new{
    Comparison of iFlow~\cite{urain2020imitationflow}, NODE~\cite{auddy2022continual} and \snode{} on LASA 2D. Results for iFlow and NODE are reproduced from~\cite{auddy2022continual}. Both NODE and \snode{} utilize a clock signal. \snode{} is comparable to NODE and is more accurate than iFlow. Note that the x-axis is log-scaled.
    }}
    \label{fig:iflow_node_snode}
\end{figure}

\new{We also compare the clock-augmented \snode{} and NODE with Imitation Flow (iFlow)~\cite{urain2020imitationflow}. Previously,~\cite{auddy2022continual} showed that NODE is preferable to iFlow due to its higher accuracy and better computational efficiency. 
\myfigure{fig:iflow_node_snode} shows that \snode{} is comparable to NODE and achieves lower prediction errors than iFlow. Due to the additional computation required to enforce stability, \snode{} needs approximately 20-25 minutes to learn a single LASA 2D task to convergence, whereas NODE requires 10 minutes on our setup. In contrast, iFlow requires approximately 60 minutes for the same number of training iterations.
Note that our proposed approach to continual learning from demonstration is not dependent
on a particular choice of dynamics model (as long as it is neural network-based). We utilize \snode{} in this paper because it exhibits better empirical performance than iFlow.
However, \snode{} can be easily swapped with a different dynamics model if so desired.}

\begin{figure}[b!]
  \includegraphics[width=0.75\textwidth]{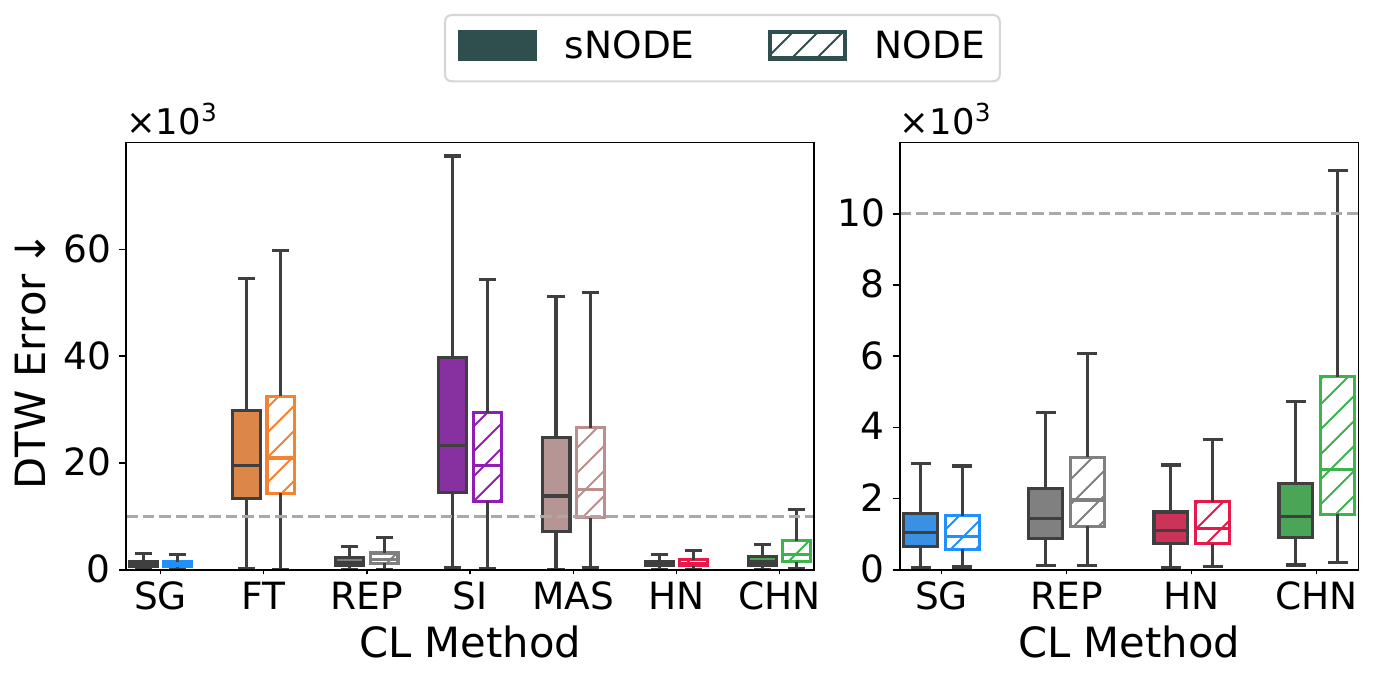}
  \caption[]{
    Overall DTW errors on LASA 2D for all CL methods~(left), and a zoomed view of the best methods~(right). The dashed gray line is a reference to compare the two plots. 
    \hnsnode{} and \chnsnode{} outperform other CL methods and perform on par with the upper baselines SG and REP.
    Stability of \snode{} improves CL performance of REP, HN and particularly that of the smallest model CHN.
    Results are obtained with 5 independent seeds
      }
      \label{fig:exp_lasa2dbox}
\end{figure}

\begin{figure}[t!]
  \includegraphics[width=0.6\textwidth]{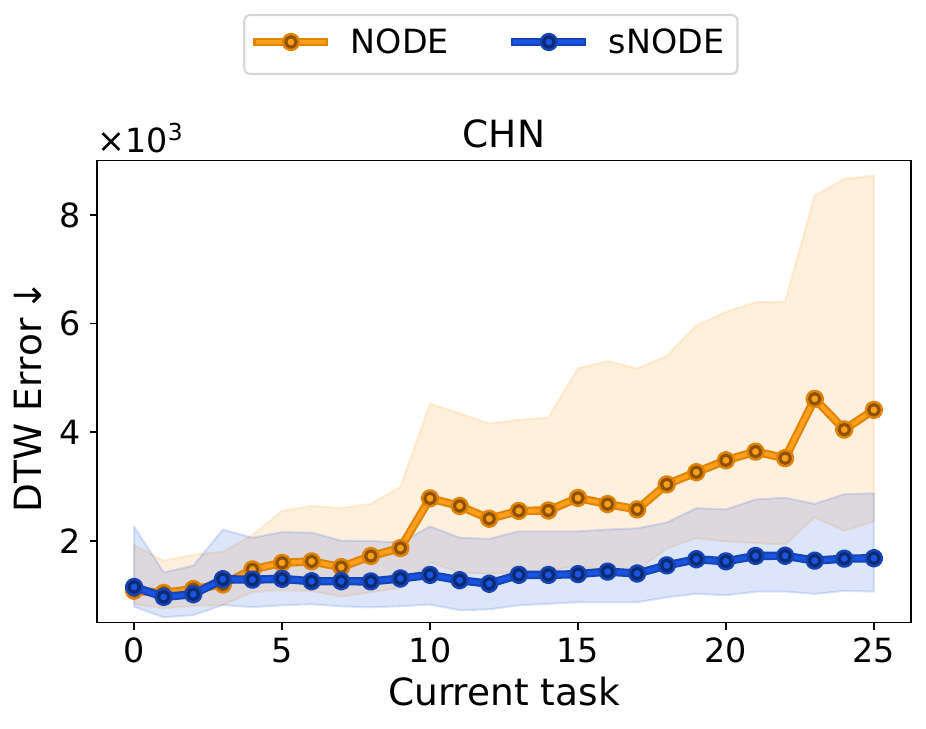}
  \caption{\chnsnode{} remembers all 26 tasks of LASA 2D, while \chnnode{} starts producing high errors after task 9. The x-axis shows the current task and the y-axis shows the errors on the current and all previous tasks after learning each task. Points show medians and shaded regions show the IQR of results for 5 independent seeds.}
  \label{fig:exp_lasa2d_cumu}
\end{figure}
\begin{figure}[t!]
  \centering
  \subfloat[]{\includegraphics[width=0.35\columnwidth]{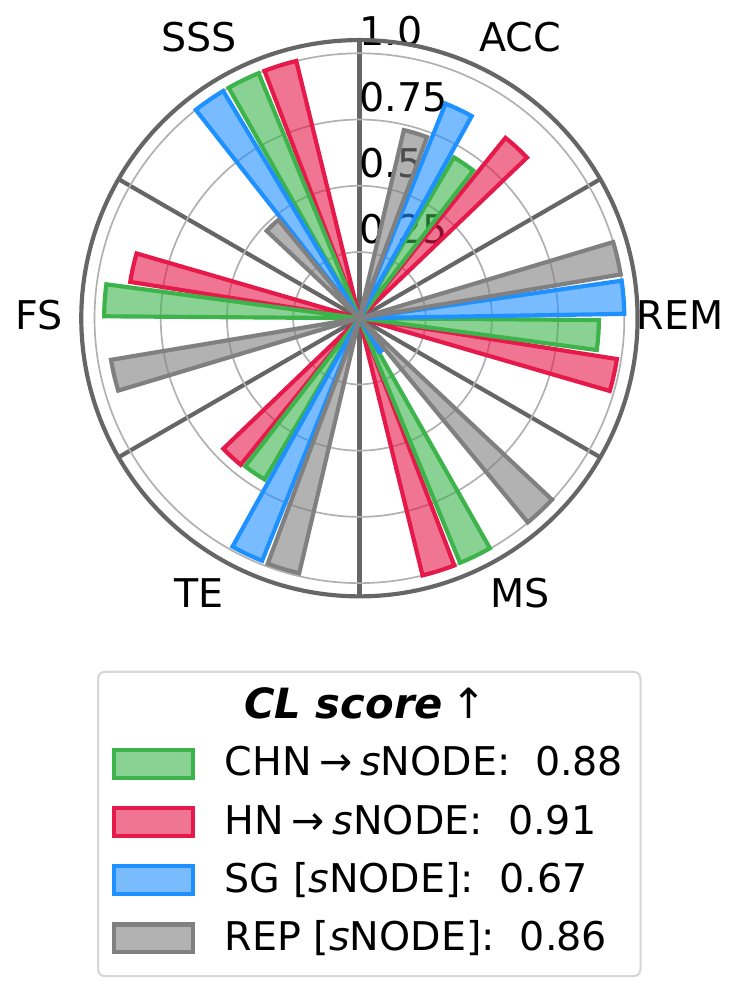}
  \label{fig:lasa2d_cl_inter_snode}}
  \subfloat[]{\includegraphics[width=0.35\columnwidth]{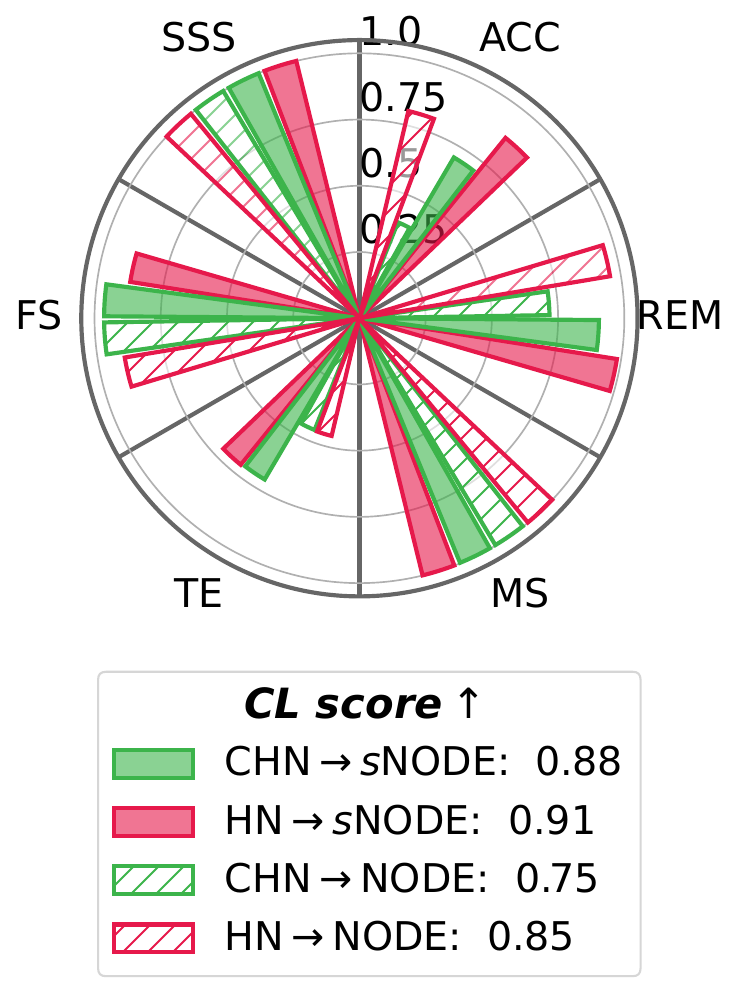}
  \label{fig:lasa2d_cl_chn_hn}}
  \caption[]{
    CL metrics (0:worst-1:best) on LASA 2D. Overall CL score is shown in the legend. 
    (a)~Comparison of CL methods using \snode{}: HN and CHN perform consistently across all CL metrics unlike SG and REP; \hnsnode{} achieves best CL score. 
    (b)~Comparison of NODE and \snode{}: \chnsnode{} outperforms \chnnode{} on multiple CL metrics such as ACC, REM and TE.
  }
  \label{fig:lasa2d_cl}    
\end{figure}

\subsection{Continual LfD on a long task sequence}
\label{sec:results_lasa_2d}

We consider the 7 types of CL methods (SG, FT, REP, SI, MAS, HN, CHN) described in \mysection{sec:baselines}, and for each CL method, we consider 2 kinds of task learning approaches: NODE and \snode{}. Thus, in total we have 14 different models, each of which we continually train on the 26 tasks of \lasa{2}. 
After a model has finished learning a task, we evaluate the prediction errors for the current task as well as all previous tasks. For example, after learning the $m^\upth$ task in a sequence of $T$ tasks, a model is evaluated on tasks ($0,1,\cdots,m$). This is repeated for all $T$ tasks and
each experiment is repeated with 5 independent seeds. 

The overall errors reported in \myfigure{fig:exp_lasa2dbox} show that amongst all CL methods, FT, SI, and MAS produce high errors (for both NODE and \snode{}), and the hypernetwork methods (HN, CHN) with \snode{} perform on par with the upper baselines SG and REP.
For SG, NODE and \snode{} have almost no difference. However, for all the other CL methods that produce low errors (REP, HN, CHN), the errors for \snode{} are lower than those for NODE. This difference is most prominent for CHN (the smallest model, see~supplementary materials for model sizes) where \chnsnode{} clearly outperforms \chnnode{}.
\myfigure{fig:exp_lasa2d_cumu} shows the evolution of prediction errors during the training process for CHN. 
\chnsnode{} remembers all 26 tasks of \lasa{2}, while \chnnode{} only remembers a few tasks and produces high errors after task 9.
Overall, \chnsnode{} performs comparably to HN and the upper baseline SG, both of which have many more parameters.%

\begin{figure}[t!]
  \includegraphics[width=0.7\textwidth]{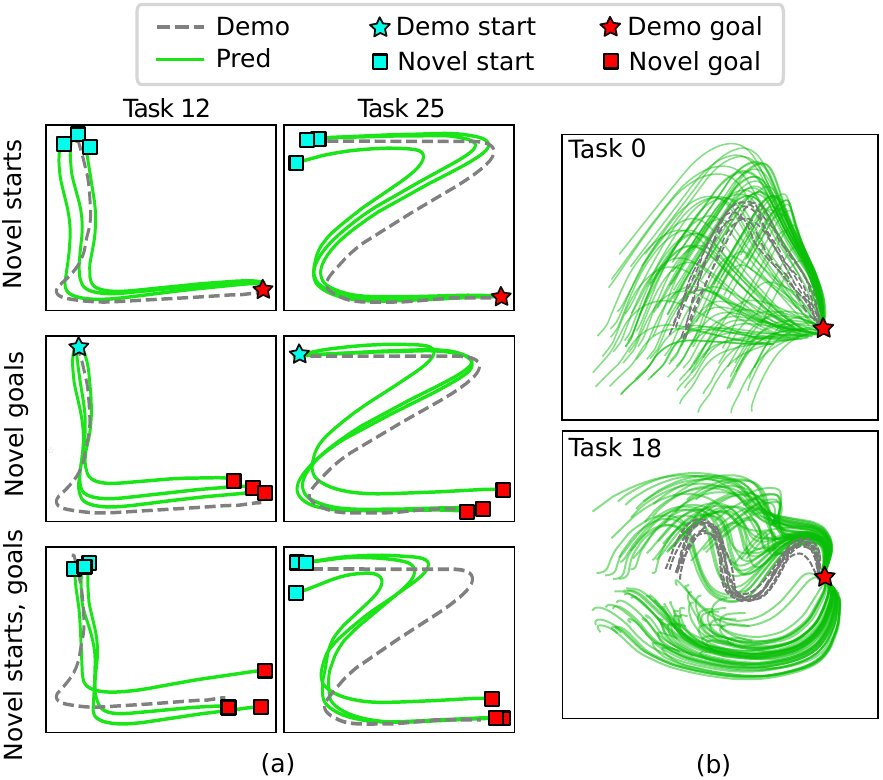}
  \caption[]{Qualitative examples for \chnsnode{} on LASA~2D.
  (a)~\chnsnode{} maintains similarity to demonstration for novel starts (1st row), novel goals (2nd row), and both novel starts and goals (3rd row), where novel starts/goals are not demonstrated.
  (b)~When novel starts are very different from demonstrations, \chnsnode{} still converges near the goal safely. However, this extreme out-of-distribution scenario requires new demonstrations to be collected.
  }
  \label{fig:qual_lasa}
\end{figure}

\begin{figure}[b!]
  \includegraphics[width=0.7\textwidth]{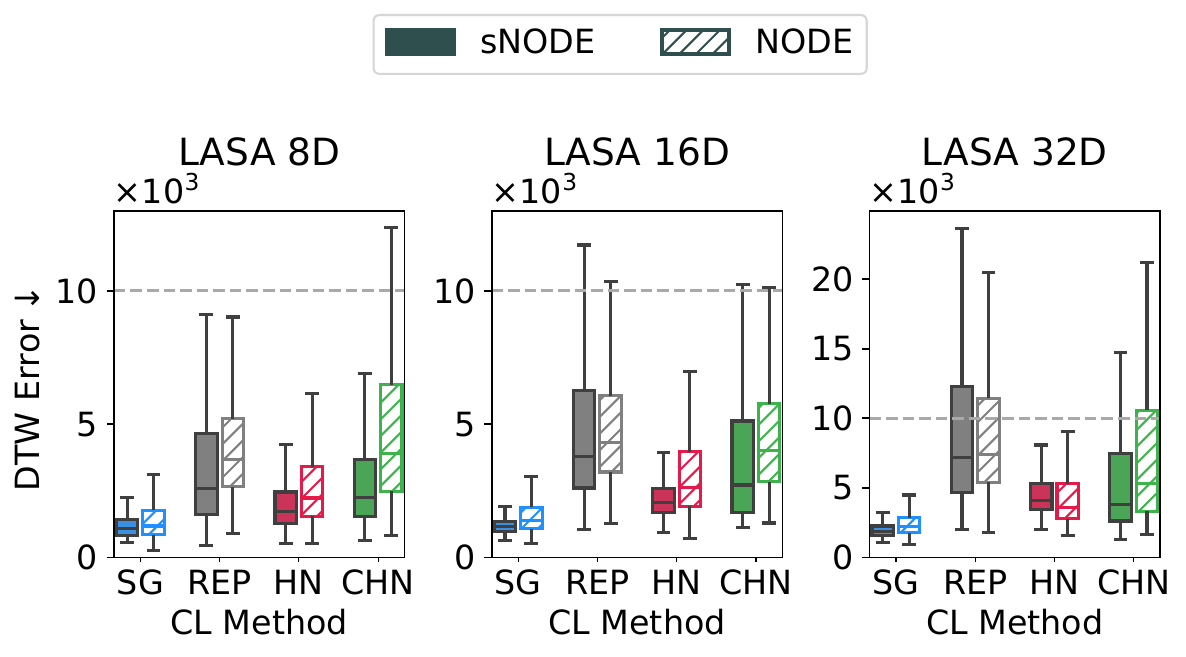}
  \caption{Comparison of overall DTW errors (y-axis) on high-dimensional LASA datasets.
  The dotted gray line is a reference for comparison. In most cases,
  \snode{} improves the performance of CHN and HN. Overall, \hnsnode{} performs closest to the upper baseline SG. Results are obtained with 5 independent seeds.}
  \label{fig:lasa_highd_box}
\end{figure}

\begin{figure}[t!]
	\includegraphics[width=0.7\textwidth]{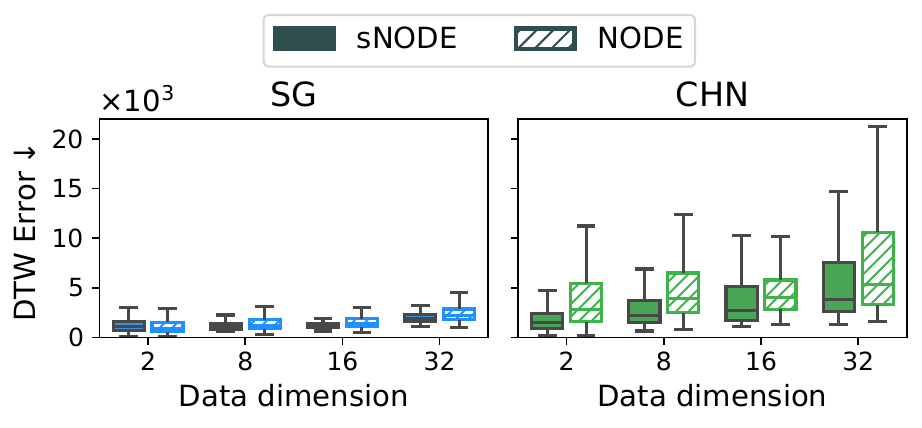}
	\caption{Comparison of overall DTW errors across all LASA datasets. Errors for CHN increase with data dimensionality, but \chnsnode{} degrades less than \chnnode{}. The upper baseline SG is shown for reference.}
	\label{fig:lasa2d_chn_dims}
\end{figure}

\begin{figure*}[t]
  \includegraphics[width=0.68\textwidth]{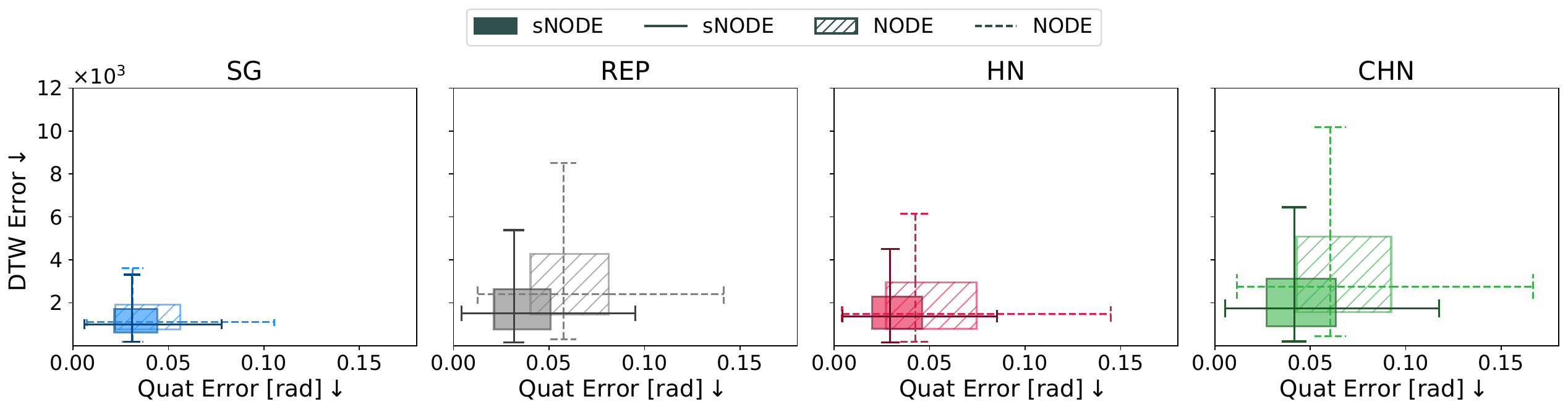}
  \caption[]{
  \snode{} improves CL performance on RoboTasks9.
  2D boxplots compare the NODE and \snode{} models of upper baselines SG and REP with HN and CHN. The overall position DTW error (y-axis) and orientation quaternion error (x-axis) of all predictions during training are shown.
  \chnsnode{} and \hnsnode{} are comparable to the upper baselines in spite of using a single model (unlike SG) and not retraining on past demonstrations (unlike REP).
  SG has almost the same performance for \snode{} and NODE with minor improvement in quat error.
  CHN improves the most with \snode{}.}
  \label{fig:robottasks_box}
\end{figure*}

Next, we report CL metrics which measure the trade-offs made by the models for learning all the tasks of \lasa{2}. To compute these CL metrics, each prediction is classified as either accurate or inaccurate by setting a threshold on the DTW error. For this, we use the same threshold value of 2191 used by \cite{auddy2022continual}. 
The resulting CL metrics for SG, REP, HN and CHN are reported in~\myfigure{fig:lasa2d_cl}.
We do not report CL metrics for SI, FT and MAS due to their high prediction errors in \myfigure{fig:exp_lasa2dbox}.
\myfigure{fig:lasa2d_cl_inter_snode} shows that with \snode{},
HN and CHN perform consistently well on all CL metrics, unlike SG and REP. 
\myfigure{fig:lasa2d_cl_chn_hn} shows that \chnsnode{} outperforms \chnnode{} on multiple metrics. 
\snode{} also improves the performance of HN, and
\hnsnode{} achieves the best overall CL score.

Qualitative examples in \myfigure{fig:qual_lasa}a show that \chnsnode{} generalizes to user-specified novel initial positions and goals not present in the training demonstrations while predicting trajectories that still resemble the demonstrations.
\myfigure{fig:qual_lasa}b highlights the stable nature of \chnsnode{} and shows that it converges near the goal even when starting from initial positions far away from what was demonstrated. Of course, when initial states are very different from demonstrations, it is not possible for predictions to resemble the demonstrated shape and new demonstrations need to be collected. 
Further empirical stability tests are reported in the supplementary materials.

\subsection{Continual LfD on high-dimensional tasks}
\label{sec:results_lasa_high-d}

The high-dimensional LASA datasets (\lasa{8}, \lasa{16}, \lasa{32}) are designed to test scalability on high-dimensional LfD tasks.
For each of these datasets, we repeat the same experiments as for \lasa{2}. We evaluate the CL methods SG, FT, REP, HN, and CHN, and consider two versions of each method, corresponding to NODE and \snode{}. We omit SI and MAS due to their poor performance on \lasa{2}, but include FT as a lower performance baseline for reference. All experiments are repeated 5 times with independent seeds.

\myfigure{fig:lasa_highd_box} shows the overall prediction errors for the current and past tasks across all tasks of each high-dimensional LASA dataset during training. FT achieves median DTW values ranging from 56$\times 10^3$ to 151$\times 10^3$ on the high-dimensional datasets and is not shown in the plots due to the large scale of these errors.
The performance of the upper baseline SG with NODE and \snode{} is similar, but \snode{} improves performance for our proposed CL methods CHN and HN in most cases. The REP baseline performs worse than HN/CHN$\rightarrow$\snode{}.
Overall, \hnsnode{} produces the lowest errors among the CL methods, performing close to the upper baseline SG. HN and CHN use a single model for learning all tasks (unlike SG, which uses a separate model for each task) and do not retrain on past tasks (unlike REP).
These benefits are also reflected in the CL metrics of the hypernetwork models reported in the supplementary materials, where \hnsnode{}, \chnsnode{} exhibit the best overall CL scores.
Among the hypernetworks, the superior performance of \hnsnode{} compared to \chnsnode{} can be attributed to its relatively larger size that enhances its representational capacity and aids performance on these high-dimensional tasks.

With an increase in the trajectory dimension, the difficulty for CL also increases. To analyze performance changes with increasing dimension, we plot the overall errors for CHN and the upper baseline SG for all LASA datasets (including \lasa{2}) side by side in \myfigure{fig:lasa2d_chn_dims}, which shows that for both NODE and \snode{}, the errors produced by CHN increase as the trajectory dimension increases from 2 to 32. However, the performance degradation is more severe for NODE than for \snode{}. 

\begin{figure}[t!]
  \centering
  \subfloat[]{\includegraphics[width=0.35\columnwidth]{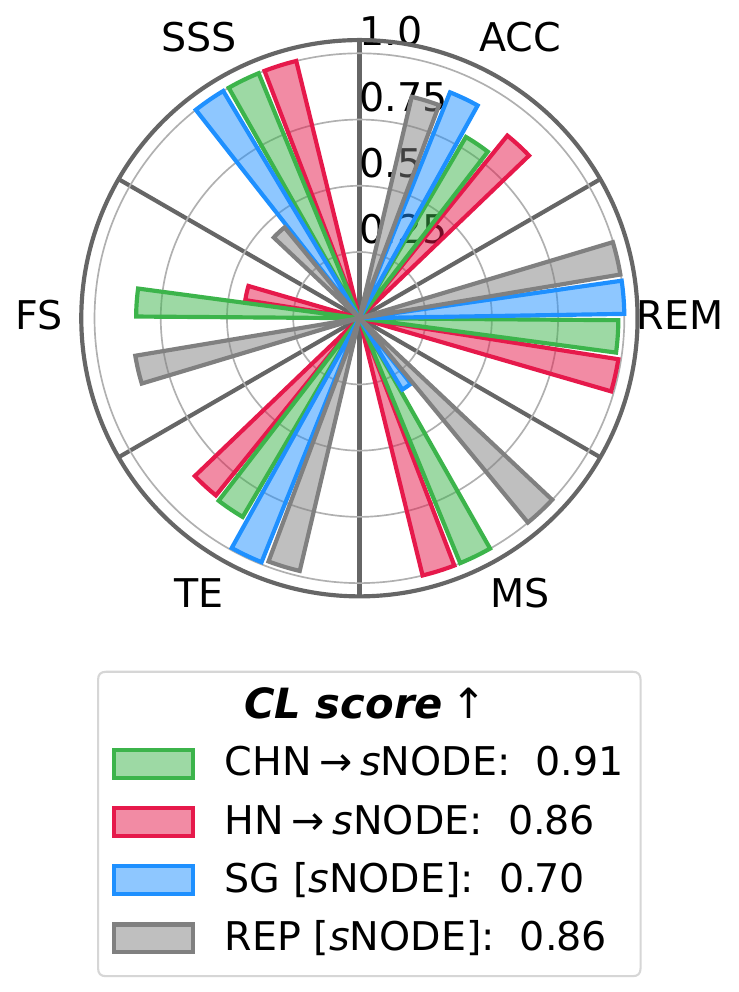}
  \label{fig:rt_pizza_a}}
  \subfloat[]{\includegraphics[width=0.35\columnwidth]{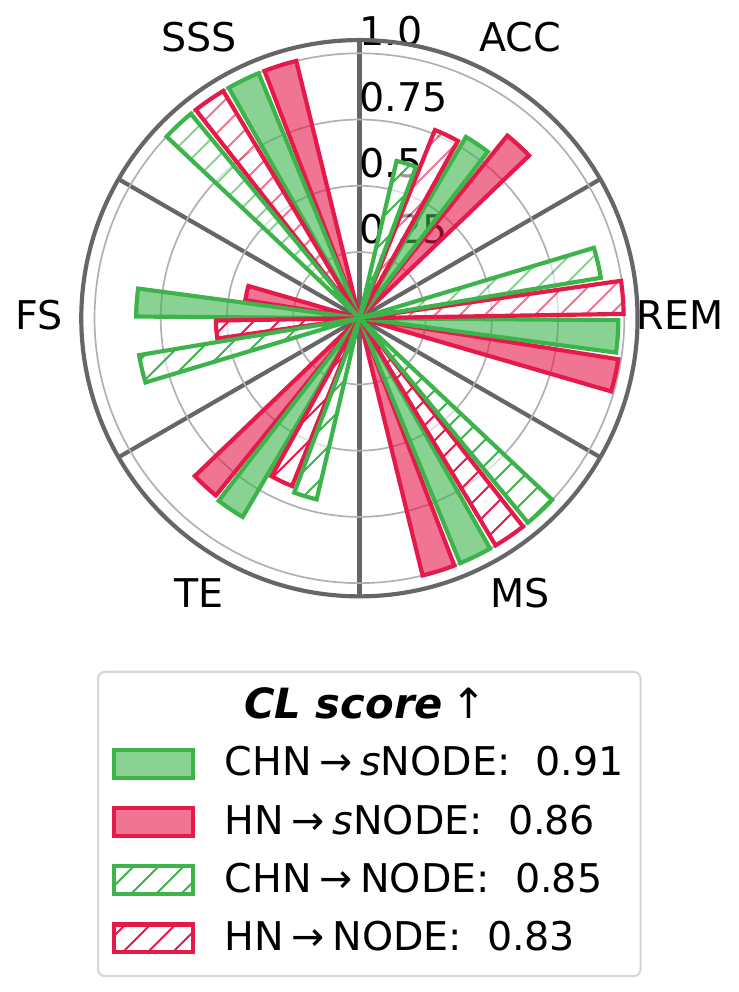}
  \label{fig:rt_pizza_b}}
  \caption[]{
    CL metrics (0:worst-1:best) for RoboTasks9. Metrics are averaged over position and orientation. Overall CL score is shown in the legend. 
    (a)~Comparison of CL methods using \snode{}: CHN achieves the best CL score by performing consistently across all CL metrics unlike HN, SG and REP.
    (b)~Comparison of NODE and \snode{} for hypernetworks. \chnsnode{} is the only method that performs consistently across all CL metrics.
  }
  \label{fig:rt_pizza}    
\end{figure}

\subsection{Continual LfD on real-world tasks}
\label{sec:results_rt}

We evaluate our approach on the sequence of 9 real-world LfD tasks in \robot{}. In each task, both position and orientation of the end-effector are learned simultaneously. We compare the CL methods SG, FT, REP, HN and CHN. Two versions of each method (corresponding to NODE~\cite{auddy2022continual} and \snode{}) are tested.
We train the 10 different models continually on the 9 tasks of \robot{}. After each task is learned, we evaluate on the newly learned task, as well as on all previous tasks. All experiments are repeated 5 times with independent seeds. 
In \myfigure{fig:robottasks_box}, we compare the overall errors. FT is excluded due to its high errors (median DTW errors for FT lie between 29$\times 10^3$ and 35$\times 10^3$, and median Quaternion errors lie between 0.26 and 0.33 radians).

\myfigure{fig:robottasks_box} shows that the position (DTW) and orientation (Quaternion) errors of \chnsnode{} and \hnsnode{} are close to those of the upper baseline SG[\snode{}] (which uses a different model for each task) and REP[\snode{}] (which stores training data from all past tasks). Further, \snode{} improves the performance of REP, HN and CHN in both DTW and Quat. errors, while SG[\snode{}] performs almost identically to SG[NODE] (minor improvement in Quaternion error). Overall, the performance of CHN improves the most when stability is introduced with \snode{}.

Next, we evaluate CL performance on \robot{}. We empirically set thresholds of 3000 on the DTW position error, and 0.08 radians ($\approx 5$ degrees) on the orientation error. These thresholds are stricter than the ones used by \cite{auddy2022continual} (DTW threshold=7191, orientation error threshold=10 degrees), and set a higher bar for evaluating CL performance. 
We first compute the CL metrics for position and orientation separately. Since all CL metrics lie in the range [0, 1], we report consolidated CL metrics for \robot{} in~\myfigure{fig:rt_pizza} by averaging each CL metric over position and orientation.
\myfigure{fig:rt_pizza_a} shows that among the \snode{} models, \chnsnode{} is the only method with high scores for all CL metrics.
In \myfigure{fig:rt_pizza_b}, it can be seen that the \snode{} models of both HN and CHN outperform the NODE models, with \chnsnode{} exhibiting good performance in key metrics such as ACC (\emph{accuracy}), REM (\emph{remembering}), and FS (\emph{final model size}). resulting in the best overall CL score among all methods.

Overall, \chnsnode{} offers the best trade-off for CL among the methods we evaluate: its size is relatively small compared to SG and HN (sizes are reported in supplementary materials), it can learn and remember multiple tasks without forgetting (see \myfigure{fig:robottasks_box}), it does not need to store and retrain on demonstrations of past tasks like REP, and it predicts stable motion, which also helps to improve its CL performance over the NODE-based alternative (see~\myfigure{fig:rt_pizza_b}). 
Fig. 6 in the supplementary materials shows some qualitative examples of the predictions for \robot{}, where it can be seen than \chnsnode{} produces low errors. The robot can be seen performing the tasks of \robot{} in the supplementary video.%

\subsection{Stochastic Hypernetwork Regularization}
\label{sec:results_stoch_reg}

To overcome the quadratic increase in hypernetwork training time with each new task, we use only a single randomly sampled past task embedding for regularization (as proposed in \mysection{sec:method_hn_sampled_reg}). 
To compare different regularization approaches, we train three types of \chnsnode{} on \robot{}, differentiated by the way regularization is performed:

\begin{enumerate}[label=(\roman*)]
    \item \emph{Full regularization}: All past task embeddings are considered for regularization~\cite{von2019continual}, as per \myequation{eq:hn_loss_step2}. This is the same as the CHN models used in all the previous experiments. We refer to this model as \emph{CHN-all}.
    \item \emph{Set-based regularization}: Task embeddings of a fixed number of randomly selected past tasks are considered for regularization~\cite{von2019continual}, as per \myequation{eq:hn_loss_step2_fix1}. We consider 2 versions of this model: \emph{CHN-3} and \emph{CHN-5}, which use 3 and 5 randomly selected past task embeddings for regularization respectively.
    \item \emph{Single task embedding}: In each training iteration, a single task embedding is uniformly sampled from the list of past task embeddings and used for regularization, as per \myequation{eq:hn_loss_step2_fix2} proposed by us. We refer to this model as \emph{CHN-1}. 
\end{enumerate} 

\begin{figure}[b!]
  \includegraphics[width=0.7\textwidth]{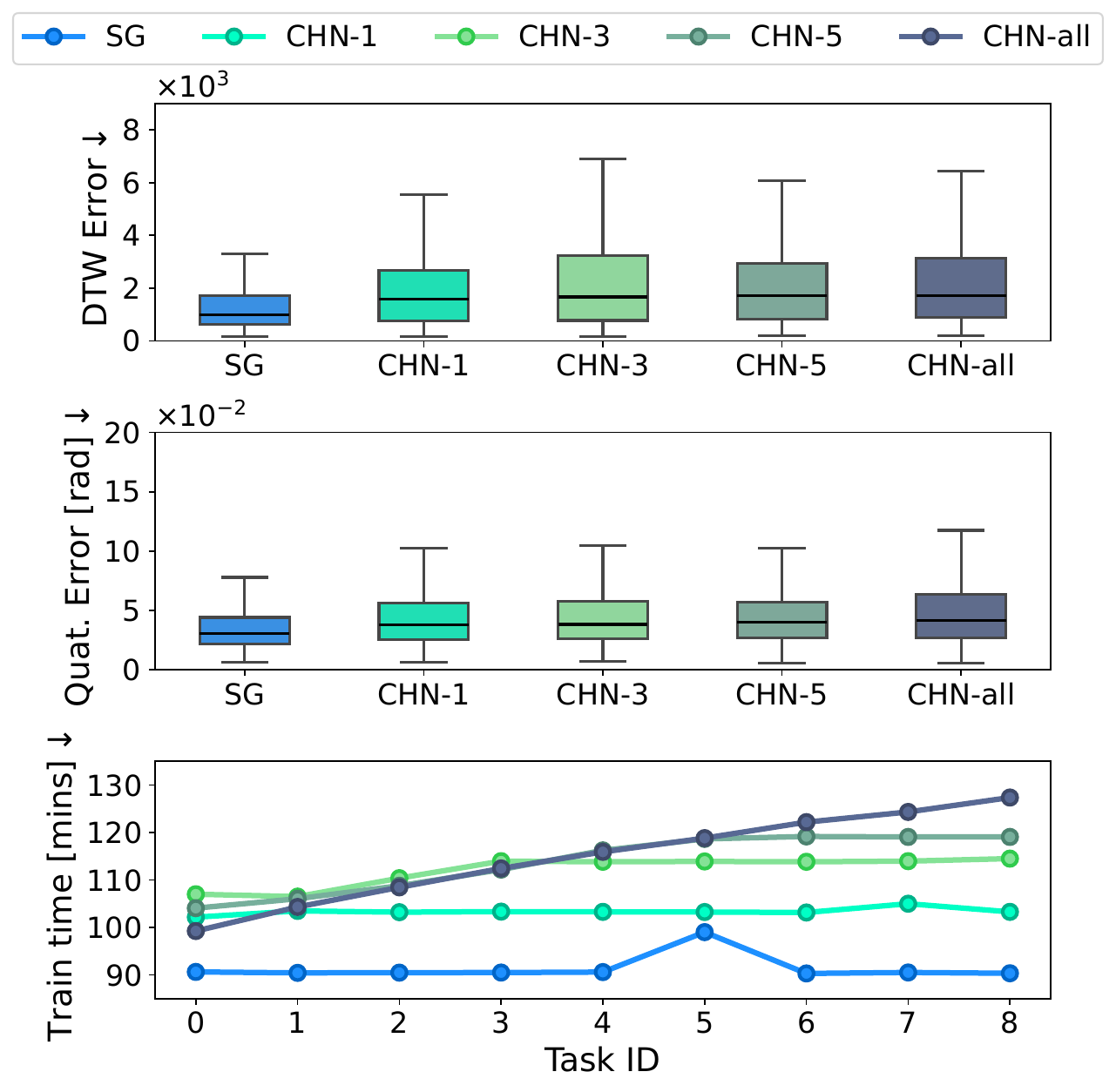}
  \caption{
    Stochastic regularization in \chnsnode{} on RoboTasks9: 
    (top)~position errors, (middle)~orientation errors, and (bottom)~training time per task. 
    Upper baseline SG (using \snode{}) is shown as a reference for good performance, but uses a different model for each task. CHN-1 (ours), CHN-3, and CHN-5 use 1, 3 and 5 randomly selected task embeddings for regularization respectively. CHN-all uses all available task embeddings. 
    CHN-1 performs equivalently to others, but its cumulative training cost is $\mathcal{O}(N)$ for $N$ tasks. The cumulative cost for CHN-3, and CHN-5 increases quadratically till 3 and 5 tasks are learned respectively. CHN-all has a cumulative cost of $\mathcal{O}(N^2)$.
  }
  \label{fig:res_rt_stoc_regu_main}
\end{figure}

We train all 4 models (CHN-all, CHN-5, CHN-3, and CHN-1) on the 9 tasks of \robot{}, and repeat the experiment 5 times with independent seeds. We use the same hyperparameters (see Tab. IV in the supp. materials) that were used for previous experiments in \mysection{sec:results_rt}. 
After each task is learned, we evaluate each model on the current task as well as all past tasks and repeat this process for all tasks in the sequence. \myfigure{fig:res_rt_stoc_regu_main} (top and middle) shows the prediction errors during this evaluation. For reference, we also show the performance of the upper baseline SG in the same plot. In \myfigure{fig:res_rt_stoc_regu_main} (bottom), we show the training time for learning each task. The number of training iterations is same for each task and all models.

In \myfigure{fig:res_rt_stoc_regu_main} (top and middle), CHN-1 performs as well as the other CHN models and the upper baseline SG. Performance is not impacted by the number of task embeddings used for regularization. 
However, the use of a single task embedding for regularization enables CHN-1 to achieve $\mathcal{O}(N)$ growth in the cumulative training time for $N$ tasks compared to the $\mathcal{O}(N^2)$ growth for CHN-all, as shown in \myfigure{fig:res_rt_stoc_regu_main} (bottom). CHN-3 and CHN-5 also have $\mathcal{O}(N^2)$ growth till 3 and 5 tasks are learned, respectively. SG is marginally better than CHN-1, but it uses a separate model for each task resulting in a much larger overall parameter size (see supplementary materials) and much worse overall CL performance (see \myfigure{fig:rt_pizza_a}).
Note that all CHN models have the same hyperparameters, are trained for the same number of iterations on each task and do not access data from past tasks.

We also train CHN-1 on all LASA datasets (2-, 8-, 16- and 32-dimensional trajectories) with the same hyperparameters used in Secs. \ref{sec:results_lasa_2d} and \ref{sec:results_lasa_high-d}. The results are presented in Fig.~5 in the supplementary materials, where CHN-1 is compared against CHN-all and SG (both with \snode{}). 
The median errors of CHN-1 are comparable to CHN-all. However, the variability for CHN-1 is much larger than CHN-all for \lasa{2}, \lasa{16}, and \lasa{32}. On \lasa{8}, CHN-1 is equivalent to CHN-all (similar to \robot{}).

CHN-1 performs as well as CHN-all on \robot{} (9 tasks, 6-dimensional trajectories)  and \lasa{8} (10 tasks, 8-dimensional trajectories), but exhibits reduced performance compared to CHN-all when the number of tasks is high (\lasa{2}), or when each task involves high-dimensional trajectories (\lasa{16}, \lasa{32}).
\new{
The remembering capacity of the hypernetwork is dependent on the regularization strength, and for CHN-1, this depends on the number of times each task is sampled for regularization.
}
Thus if the number of tasks is high (as in \lasa{2}), it is possible that each task embedding is not sampled frequently enough to remember past tasks well. 
Also, if the number of training iterations is low compared to the task complexity (in terms of dimensions), CHN-1 can also suffer due to insufficient regularization. 
For \lasa{2}, we use $1.5\times 10^4$ iterations per task (based on~\cite{auddy2022continual}), and for \robot{} ($4\times 10^4$ iterations per task) and \lasa{8} ($6\times 10^4$ iterations/task), we scale the iterations approximately linearly based on the trajectory dimensions. 
However, for the higher-dimensional \lasa{16} ($7\times 10^4$ iterations/task) and \lasa{32} ($8\times 10^4$ iterations/task) datasets, we used much fewer iterations than that suggested by this dimension-based proportional scaling to limit the overall run time of our many experiments.
To make CHN-1 more effective, a naive solution can be to simply increase the number of training iterations. A better approach can be to use a priority-based sampling of past task embeddings during regularization such that important task embeddings are sampled more frequently. 
\newsecond{
Note that all hyperparameters and network architectures are reported in Tab.~IV in the supplementary materials.
}

\newsecond{
As practical guidance, we recommend CHN-1 when the number of tasks to remember is around 10 and the trajectory dimension does not exceed 10. For more complex or numerous tasks, or when the training time is not a constraint, the full regularization is preferable. During evaluation, the overall error (DTW, Quat. error) across past tasks is an indicator of regularization effectiveness. When CHN-1's sparse regularization underperforms but training time must still be minimized, the set-based regularization strategy is a viable alternative. In this case, we recommend starting with a smaller regularization set and gradually increasing its size until the desired balance between performance and training efficiency is reached.
}

Overall, the stochastic regularization process of CHN-1 proposed in this paper is an effective strategy for continually learning real-world LfD tasks (\robot{}) or tasks of a similar nature (\lasa{8}). For these situations, the cumulative training cost is reduced to $\mathcal{O}(N)$ from $\mathcal{O}(N^2)$ without any loss in performance. 
\new{
}
In the future, we will investigate approaches to overcome the current limitations of CHN-1 to make it an effective strategy for more complex scenarios involving a higher number of tasks and/or high-dimensional trajectories. 

\subsection[short]{\new{Empirical Stability Tests}}
\label{sec:stability}

\begin{figure}[t!]
  \centering
  \subfloat[]{\includegraphics[width=0.5\columnwidth]{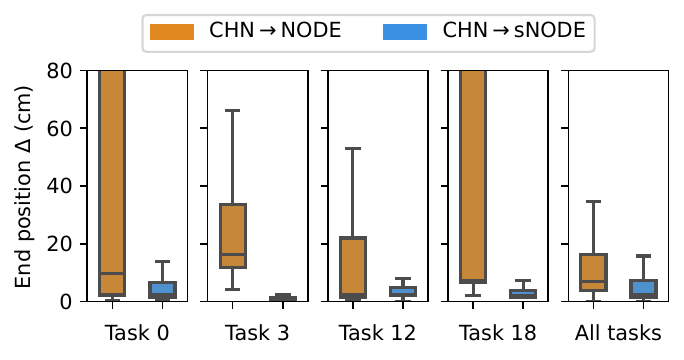}
  \label{fig:stab_box_a}}
  \subfloat[]{\includegraphics[width=0.4\columnwidth]{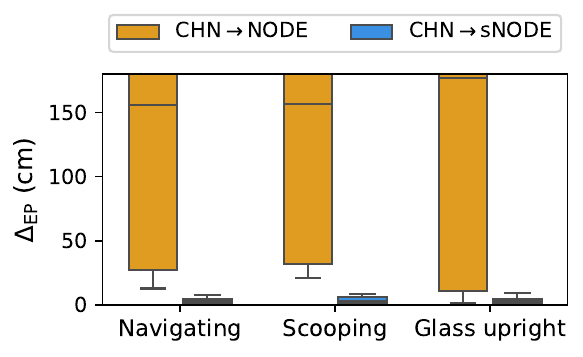}
  \label{fig:stab_box_b}}
  \caption[]{
    Quantitative results for stability test of \chnnode{} and \chnsnode{} on (a) \lasa{2}, and (b) \robot{}. For each task, the starting position is set randomly around the demonstration starting position, and we measure \new{$\Delta_\text{EP}$}, the distance between the demonstration goal and the end point of the predicted trajectory. \chnnode{} shows divergence for multiple tasks, but \chnsnode{} converges near or at the goal.
  }
  \label{fig:stab_box}    
\end{figure}

\new{We  test the stability of trajectories predicted by our proposed continual LfD models on \lasa{2} and the real-world LfD tasks in \robot{}.}
For \lasa{2}, we initialize trained models of \chnnode{} and \chnsnode{} (after learning the 26 tasks of the dataset) with starting positions that are different from the demonstrations. For each model and each past task, we choose 100 random starting positions by uniformly sampling points that lie within a box of side 50 cm centered at the ground truth start position. Qualitative examples of the predicted trajectories for a couple of tasks were presented earlier in \myfigure{fig:qual_lasa}, and here we perform the quantitative evaluation. For each random start position, we measure \new{$\Delta_\text{EP}$, the \emph{difference} between the \emph{end point} of the predicted trajectory and the demonstration goal. $\Delta_\text{EP}$ is defined as}:
\new{
\begin{equation}
    \Delta_\text{EP} = \vert\vert \hat{\mathbf{p}}_{T-1} - \mathbf{p}_{T-1}\vert\vert_2 \label{eq:ep}
\end{equation}
where $\mathbf{p}_{T-1}$ is the ground truth goal, $\hat{\mathbf{p}}_{T-1}$ is the predicted goal, and $\hat{\mathbf{p}}_{0} = \mathbf{p}_{0} + \epsilon \sim U(-l,l)$ is a random initial position.
}
\myfigure{fig:stab_box_a} shows that \chnnode{} produces divergent trajectories for multiple tasks, while \chnsnode{} does not, leading to lower values of \new{$\Delta_\text{EP}$}.
The large number of tasks in \lasa{2} allows us to test the convergence/divergence of the predictions for a variety of trajectory shapes and start positions. 
\new{
\myfigure{fig:stab_box_b} shows the quantitative results for the same test performed for some of the \robot{} tasks. Here, 90 random initial positions where sampled uniformly from within a box of side 20 cm, centered at the ground truth start position. \chnnode{} shows severe divergence, but \chnsnode{} converges near the goal despite starting from novel initial conditions. This is also demonstrated by the qualitative examples of the trajectories shown in~Fig.~2 in the supplementary materials. Further stability tests can also be found in the supplementary materials.
}
\newer{Additionally, we show that it is possible to learn from noisy demonstrations (see Sec. VI in the supplementary materials).}

\new{
Note that \chnnode{} and \chnsnode{} generate trajectories for the robot to execute, and with these tests, we evaluate how robust and accurate the trajectory generation is against novel conditions that are very different to the demonstrations. In this work, we treat trajectory generation and trajectory execution separately. Similar to \cite{otto2023deep, celik2024acquiring, celik2022specializing}, our method is responsible for generating trajectories that we assume the robot's motion controller can execute accurately while handling disturbances experienced during motion execution.
}

\section{Discussion}
\label{sec:discussion}

We proposed hypernetworks that generate stable dynamics models as an approach to stable, continual LfD. For learning each LfD task we followed the approach of simultaneously learning trajectories and stability with dedicated neural networks~\cite{kolter2019learning,lawrence2020almost}.
We proposed two types of hypernetworks (\hnsnode{}, \chnsnode{}), where each model learns multiple LfD tasks continually.
We showed that stability in our continual LfD system enhances CL performance (i.e., the ability to remember multiple LfD tasks without forgetting). 
\new{
Through experiments on several LfD datasets, we present empirical evidence that the stability of our proposed \chnsnode{}/\hnsnode{} improves continual learning performance on long sequences of tasks~(\myfigure{fig:exp_lasa2d_cumu}, \myfigure{fig:lasa2d_cl}(b)), sequences of high-dimensional LfD tasks~(\myfigure{fig:lasa_highd_box}), as well as sequences of real-world LfD tasks~(\myfigure{fig:robottasks_box}, \myfigure{fig:rt_pizza}).
The inductive bias of stability results in a model that always converges to the goal.
Even a randomly initialized \chnsnode{} (without being trained) predicts trajectories that converge at the goal (see Fig.~4 in the supplementary materials). During training, all it needs to learn is to replicate the trajectory shapes of the demonstrations.
For most tasks, the final approach to the goal can be approximately linear, and here the \chnsnode{}/\hnsnode{} does not need to expend parameters to learn this behavior, thereby freeing up resources that can be better utilized for continual learning. In contrast, \chnnode{}/\hnnode{} must learn to replicate the demonstrations as well as to converge at the goal, putting greater pressure on its continual learning capability.
}

The improvement in CL performance is most pronounced for the size-efficient \chnsnode{}. 
Instead of magnifying learning difficulty, the stability constraint drives \chnsnode{} to efficiently use its limited set of parameters, and makes it a viable option for continual LfD on resource-constrained platforms.
\hnsnode{} is more expressive due to its larger size, and is a good choice for continually learning high-dimensional LfD tasks.
Both models grow negligibly with new tasks and do not store or retrain on demonstrations from past tasks.
\new{We also showed that our proposed hypernetworks perform favorably compared to other CL approaches (dynamic architectures, replay, direct regularization)}.
\newer{Note that, following theoretical work~\cite{littwin2020infinite}, we keep our network architectures relatively shallow and wide to help the training of hypernetworks that generate \snode{}s (see Tab. IV. in the supplementary materials). Furthermore, the inherent stability of the \snode{} also benefits the overall training convergence, as mentioned above, and the hypernetwork training remains scalable to high-dimensions and long task sequences.}

We introduced a clock-augmented \snode{} that improves accuracy while retaining stability~(\mysection{sec:results_snode}).
We also introduced stochastic hypernetwork regularization with a \emph{single} task embedding to improve training efficiency~(\mysection{sec:results_stoch_reg}).
Finally, we proposed new datasets for evaluating LfD~(\mysection{sec:experiments_datasets}).
To aid reproducabilty and to support research in continual LfD, our open-source code and datasets will be released upon acceptance.

\smallskip

\noindent \textbf{Future Work}:
Our proposed hypernetworks, particularly \hnsnode{}, are effective at continually learning high-dimensional LfD tasks. \newer{In the future, we will add methodological advances, such as extending our current setup to include images (e.g. from a gripper-mounted camera), latent representations derived from images~\cite{sochopoulos2024learning}, point clouds~\cite{byravan2017se3}, or sequences of these high-dimensional features in the demonstrations used for training.} 
We will also investigate ways to improve \chnsnode{}'s performance on high-dimensional tasks, possibly by using an adaptive regularization constant that can achieve a better \emph{stability-plasticity} trade-off in CL. Additionally, tuning important \snode{}-specific hyperparameters (such as $\alpha$ in \myequation{eq:snode}) can be an option worth exploring. 

The neural network inside a hypernetwork model does not grow with new tasks. At some stage, over-saturation can prevent new tasks from being learned. A graceful forgetting mechanism can help here, by ignoring task embeddings from old or irrelevant tasks. 

The stochastic regularization process with a single task embedding (CHN-1) performs sub-optimally when the number of tasks is high or when the training iterations are insufficient for the complexity of the high-dimensional tasks. \newer{To address this problem, future work can utilize task embeddings in a smarter manner, e.g. using prioritized sampling of important embeddings instead of uniform sampling.}

Other ways of extending our current work may include effectively chaining together multiple tasks learned by a hypernetwork or to develop a high-level planner that leverages the diverse set of learned tasks as a versatile skill library for more complex applications.
Another possible avenue for future work can be to develop a mechanism for interpolating between learned skills using the stored task embeddings.
\newer{
Providing a theoretical analysis of the empirical effects discussed in this paper is also an important direction for future work.
}

\section{Conclusion}
\label{sec:conclusion}

We presented an approach to stable, continual LfD using hypernetwork-generated, clock-augmented \snode{}s and showed that stability improves CL performance and scalability. We also improved hypernetwork training efficiency and proposed new LfD datasets.
We reported quantitative results on trajectory errors, CL performance, and stability, and qualitative evaluations on real-world LfD tasks with a physical robot.
Our proposed hypernetworks outperform other CL methods in experiments spanning multiple datasets with varying numbers of tasks (7 to 26 tasks), trajectory dimensions (2 to 32 dimensions), and real-world LfD tasks with 6-DoF pose trajectories.  
To the best of our knowledge, this work is the first to demonstrate continual learning of high-dimensional LfD tasks, and our findings indicate that hypernetwork-generated \snode{}s are effective for continually learning both high-dimensional and real-world LfD tasks.

\balance

\bibliographystyle{IEEEtran}
\bibliography{references.bib}

\pagebreak


\setcounter{section}{0}

{\centering\Large\bfseries Supplementary Material\par}

\section{Baselines and CL Metrics}

\newcolumntype{C}[1]{>{\centering\arraybackslash}p{#1}}
\newcolumntype{L}[1]{>{\raggedright\arraybackslash}p{#1}}

\new{
We provide details of our experimental baselines in \mytable{tab:baselines}. Details and formulae of the CL metrics reported by us are provided in \mytable{tab:cl_metrics_table}. Please refer to sections V.B and V.C in the main paper for the relevant text.
}

\section{Model sizes}

\begin{figure}[h]
  \centering
  \subfloat[LASA 2D]{\includegraphics[width=0.5\columnwidth]{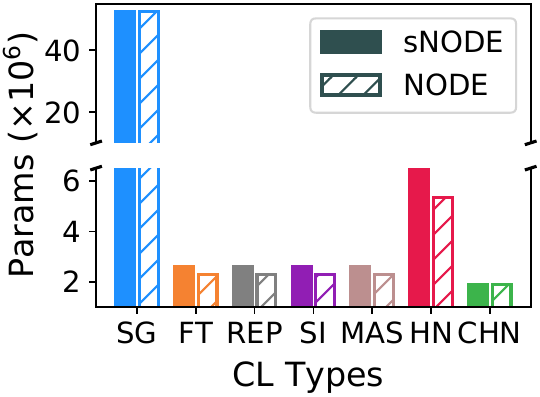}
  \label{fig:size_a}}
  \subfloat[RoboTasks9]{\includegraphics[width=0.485\columnwidth]{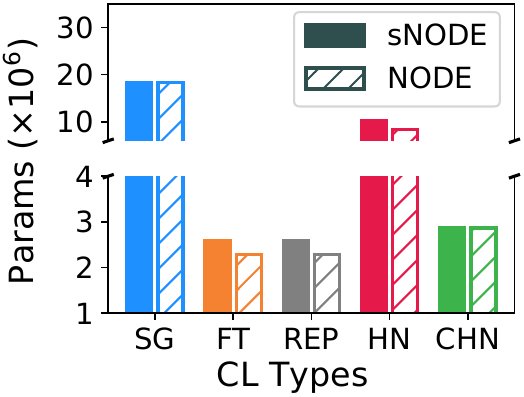}
  \label{fig:size_b}}
  \caption[]{
    Model parameter sizes after learning all tasks. We do not show the storage needed by REP to store the data of all tasks. Despite having much fewer parameters than SG and HN, \chnsnode{} performs on par with these larger models.  
  }
  \label{fig:size}    
\end{figure}

We report the sizes of continual learning (CL) models after learning all tasks in \lasa{2} and \robot{} in \myfigure{fig:size_a} and \myfigure{fig:size_b}, respectively. In all experiments, the overall size of the CL models with NODE and \snode{} are roughly equivalent. SG uses a separate network for each task, resulting in a high overall parameter count. CHN is the smallest among the compared models. Model architectures are reported in Tab. 1 in the appendix of the main paper.

\section{Complexity of Stable NODE and Hypernetworks}

\new{
\begin{claim}

If the \snode{} function $\mathbf{f}_{\upphi}(\mathbf{x}):\mathbb{R}^d \rightarrow \mathbb{R}^d$ is given by $$\mathbf{f}_{\upphi}(\mathbf{x}) = \hat{\mathbf{f}}_{\uptheta}(\mathbf{x}) - \nabla V_\upgamma(\mathbf{x}) \frac{\mathrm{ReLU}(\nabla V_\upgamma(\mathbf{x})^\newsecond{\top} \hat{\mathbf{f}}_{\uptheta}(\mathbf{x})) + \alpha V_\upgamma(\mathbf{x})}{||\nabla V_\upgamma(\mathbf{x})||^2_2},$$ the complexity of $\mathbf{f}_{\upphi}(\mathbf{x})$ is $\mathcal{O}(d)$.
\label{lemma:snode_complexity}
\end{claim}

\begin{proof}
$\mathbf{f}_{\upphi}(\mathbf{x})$ consists of the following operations:
\begin{itemize}
    \item $\hat{\mathbf{f}}_{\uptheta}(\mathbf{x})$: Assuming $\hat{\mathbf{f}}_{\uptheta}:\mathbb{R}^d \rightarrow \mathbb{R}^d$ has $H_f$ hidden layers, each with $h_f$ units. The weight matrices for this network are $W_\text{in} \in \mathbb{R}^{d \times h_f}, \{ W_i\}_{i=1}^{H_f-1} \in \mathbb{R}^{h_f \times h_f}, W_\text{out} \in \mathbb{R}^{h_f \times d}$. Ignoring the bias terms and noting that the complexity of $\sigma(\cdot)$ is $\mathcal{O}(1)$ (where $\sigma(\cdot)$ denotes the activation function, the complexity of a forward pass through  $\hat{\mathbf{f}}_{\uptheta}(\mathbf{x})$ is $\mathcal{O}(d\cdot h_f + (H_f-1)\cdot h_f\cdot h_f + h_f\cdot d)$. Generally, $d \ll h_f$ and the complexity is dominated by $h_f$. If we assume a fixed network architecture (i.e., $H_f$ and $h_f$ are constants), then the complexity of the forward pass is $\mathcal{O}(d)$.
    \item $V_\upgamma(\mathbf{x})$: Following a similar logic as above, the complexity of a forward pass through $V_\upgamma(\mathbf{x}):\mathbb{R}^d \rightarrow \mathbb{R}^1$ is also $\mathcal{O}(d)$.
    \item $\nabla V_\upgamma(\mathbf{x})$: Computation of the gradient involves similar matrix multiplications as the forward pass. In this case, the complexity is also $\mathcal{O}(d)$.
    \item $\mathrm{ReLU}(\cdot)$: This is a constant operation with complexity $\mathcal{O}(1)$.
    \item $\vert\vert\cdot\vert\vert$: The Euclidean norm involves $d$ multiplications and $d$ additions. Thus, its complexity is $\mathcal{O}(d)$.
\end{itemize}
Thus, the overall complexity of $\mathbf{f}_{\upphi}(\mathbf{x}):\mathbb{R}^d \rightarrow \mathbb{R}^d$ is $\mathcal{O}(d)$.

\end{proof}
}

\begin{table}[b]
  \caption{Consolidated CL metrics for the high-dimensional LASA datasets, computed by taking the average of each CL metric over the 3 high-dimensional LASA datasets. Values are from 0:worst to 1:best. The best value in each column is highlighted in bold. For the overall CL scores, the second-best values are underlined. sN indicates \snode{} and N indicates NODE.}
  \centering
  \resizebox{0.9\columnwidth}{!}{
      \input{LASA_highD_all.tex}
  }
  \label{tab:lasa_highd_cl}
\end{table}

\new{
\begin{claim}
    Complexity of full regularization of a hypernetwork is $\mathcal{O}(m^2)$, whereas the complexity of stochastic regularization is $\mathcal{O}(m)$, where $m$ is the number of learned tasks.
\end{claim}

\begin{proof}
    The loss in the second optimization stage for training a hypernetwork, as shown in Eq. 6 of the main paper, is given by: 
    \begin{align*}
	\tilde{\mathcal{L}}^m = &\mathcal{L}^m + \cfrac{\beta}{m-1} \sum\limits^{m-1}_{l=0}\left\vert\left\vert\mathbf{f}(\mathbf{e}^l, \mathbf{h}^*) - \mathbf{f}(\mathbf{e}^l, \mathbf{h}+\Delta\mathbf{h})\right\vert\right\vert^2
	\label{eq:hn_loss_step2}
    \end{align*}
    where $\mathcal{L}^m$ is the task specific loss, $m$ is the number of tasks, $\beta$ is the regularization constant, $\mathbf{f}(\cdot, \cdot)$ denotes the hypernetwork function, $\mathbf{h}$ denotes the hypernetwork weights, and $\mathbf{e}^l$ denotes the task embedding for the $l^\text{th}$ task.

    In each training iteration, let $C_\text{task}$ be the cost of computing the task-specific loss $\mathcal{L}^m$ and let $C_\text{reg}$ be the cost of computing the regularization cost for a single past task. Let $i$ be the number of training iterations used to learn each task. Thus, for task 0, the cost is $i \times C_\text{task}$ (since there are no past tasks before task 0), for task 1, the cost is $i \times (C_\text{task} + C_\text{reg})$, $\cdots$. In general, for $t$ past tasks, the cost is $i \times (C_\text{task} + t \times C_\text{reg})$. Therefore, for $m$ tasks, the total cost is 
    \begin{align*}
        &\sum\limits_{t=0}^{m-1} i (C_\text{task} + t C_\text{reg}) \\
        &=i \left(m C_\text{task} + C_\text{reg} \sum\limits_{t=0}^{m-1} t \right) = i \left(m C_\text{task} + C_\text{reg} \frac{(m-1)m}{2} \right)
    \end{align*}
    and so the complexity is $\mathcal{O}(m^2)$.

    The form of stochastic regularization proposed by us (Eq. 10 in the main paper) is given by
    \begin{align*}
	\tilde{\mathcal{L}}^m = &\mathcal{L}^m + \beta \left\vert\left\vert\mathbf{f}(\mathbf{e}^k, \mathbf{h}^*) - \mathbf{f}(\mathbf{e}^k, \mathbf{h}+\Delta\mathbf{h})\right\vert\right\vert^2
    \end{align*}
    where $\mathbf{e}^k$ is the task embedding sampled from the set of past task embeddings. For task 0, the cost is $iC_\text{task}$ and for each subsequent task, the cost is $i (C_\text{task} + C_\text{reg})$. Thus for $m$ tasks, the overall cost is $i \left(\sum\limits_{t=0}^{m-1}  C_\text{task} + \sum\limits_{t=1}^{m-1} C_\text{reg} \right) = i \left(m C_\text{task} + (m-1)C_\text{reg} \right)$, and thus the complexity of the stochastic regularization loss is $\mathcal{O}(m)$.
\end{proof}
}

\begin{table}[h]
    \centering
    \footnotesize
    \new{
    \begin{tabularx}{0.95\textwidth}{
        L{50pt}L{50pt}X        
     }
     \toprule 
     \textbf{Baselines} & \textbf{CL Type} & \textbf{Description} \\
     \midrule
     SG-NODE SG-\snode{} & Dynamic growth & A separate NODE or \snode{} is trained for each new task using Eq. 2 in the main paper and frozen after learning. \\ \midrule
     FT-NODE FT-\snode{} & None & A common NODE or \snode{} is trained on the first task and then sequentially finetuned on each subsequent task without any mechanism to prevent catastrophic forgetting. Eq. 2 in the main paper is used as the training loss. \\ \midrule
     REP-NODE REP-\snode{} & Replay & A common NODE or \snode{} is trained on the first task. For each subsequent task, the data from all previous tasks is combined with the data of the current task and is used to train the NODE/\snode{} with multi-task learning following the approach from~\cite{auddy2022continual}. Eq. 2 in the main paper is used as the optimization objective.\\ \midrule
     MAS-NODE MAS-\snode{} & Regularization & These baselines are similar to FT, but here, the \emph{Memory Aware Synapses} (MAS) algorithm~\cite{aljundi2018MAS} is used to prevent catastrophic forgetting. MAS uses Eq. 2 in the main paper along with a regularization loss that penalizes changes to the NODE/sNODE parameters that are important for remembering past tasks. Parameter importance is computed based on the gradient of the squared L2 norm of the output~\cite{aljundi2018MAS,auddy2022continual}. \\ \midrule
     SI-NODE SI-\snode{} & Regularization & These baselines are similar to MAS, but use a different CL algorithm known as \emph{Synaptic Intelligence} (SI)~\cite{zenke2017SI} to prevent forgetting. The regularization loss in SI is also based on parameter importance, but here a parameter's importance is related to how much it contributes to a change in the loss~\cite{zenke2017SI, auddy2022continual}. \\ \midrule
     \hnnode{} & Meta-model regularization (hypernetwork is regularized, but not the NODE) & A hypernetwork generates the parameters of a NODE for each new task~\cite{auddy2022continual}. For each task, a separate task embedding is learned and frozen afterward. The hypenetwork is shared across tasks. \\ \midrule
     \chnnode{} & Meta-model regularization & Similar to HN, but uses a \emph{chunked hypernetwork} to generate a NODE~\cite{auddy2022continual}. \\
     \midrule
     \hnsnode{} (ours) & Meta-model regularization (hypernetwork is regularized, but not the \snode{}) & A hypernetwork generates the parameters of an \snode{} for each new task. For each task, a separate task embedding is learned and frozen afterward. The hypenetwork is shared across tasks. \\ \midrule
     \chnsnode{} (ours) & Meta-model regularization & Similar to HN, but uses a \emph{chunked hypernetwork} to generate an \snode{}. \\
    \bottomrule
    \end{tabularx}
    }
    \caption{Baselines.}
    \label{tab:baselines}
\end{table}
\begin{table}[H]
    \footnotesize %
    \centering
        \new{
            \begin{tabularx}{\linewidth}{bb}
                \toprule
                \heading{Metric} & \heading{Relevance}  \\
                \midrule
                Average accuracy~\cite{diaz2018don}
                \begin{equation*}
                    \text{ACC}=\frac{\sum_{i\geq j}^NR_{i,j}}{N(N+1)/2}
                \end{equation*}
                where $R_{i,j}$ is the accuracy\footnote{Every predicted trajectory is classified as accurate or inaccurate by thresholding on its DTW distance to the ground truth trajectory.} on task $j$ after learning task $i$, and $N$ is the total number of tasks
                & \vfill Percentage of correct predictions made for the current and past tasks \\
                \midrule
                Remembering~\cite{diaz2018don}
                \begin{equation*}
                    \text{REM} = 1 - |\min(\text{BWT},0)|
                \end{equation*}
                where the backward transfer BWT is defined as $$\text{BWT}=\frac{\sum_{i=2}^N\sum_{j=1}^{i-1}(R_{i,j}-R_{j,j})}{N(N-1)/2}$$  
                & REM is a measure of forgetting and is based on the Backward Transfer (BWT) metric. BWT measures how much the accuracy on a task changes as new tasks are learned. Since BWT can be negative, REM quantifies only the forgetting part, i.e. how much the accuracy degrades.
                \\
                \midrule
                Model size efficiency~\cite{diaz2018don} 
                $$\text{MS}=\min(1,\frac{\sum_{i=1}^N\frac{\text{Mem}(\theta_1)}{\text{Mem}(\theta_i)}}{N})$$
                where $\text{Mem}(\theta_i)$ is the parameter size after learning task $i$
                & MS measures the relative growth factor in model parameters with respect to the parameter size after learning the first task.\\
                \midrule
                Samples storage size efficiency \cite{diaz2018don}
                $$\text{SSS}=1-\min(1,\frac{\sum_{i=1}^N\frac{\text{Mem}(M_i)}{\text{Mem}(D)}}{N})$$  where $\text{Mem}(M_i)$ is the storage space for task $i$ and $\text{Mem}(D)$ is the cumulative storage for all tasks.
                & SSS measures the growth in storage required to explicitly store training data from previous tasks. This is relevant for CL strategies that rely on replaying past data.\\
                \midrule
                Time Efficiency~\cite{auddy2022continual} 
                $$\mathrm{TE}=\min\left\{1,\frac{\mathcal{T}_1}{N}\sum_{i=1}^{N}\frac{1}{\mathcal{T}_i}\right\}$$ 
                where $\mathcal{T}_i$ is the time needed to learn task $i$.
                & TE measures how much the training duration increases with the number of tasks, relative to the time taken to learn the first task.\\
                \midrule
                Final Model Size~\cite{auddy2022continual}
                $$\mathrm{FS}=1-\overline{\mathrm{Mem}}(\mathrm\theta^{N})$$
                where $\overline{\mathrm{Mem}}(\mathrm\theta^{N})$ is the parameter size after learning $N$
 tasks, normalized by the size of the largest compared model.
                & FS is a relative measure of the parameter size of a model after learning all tasks compared to all evaluated models.\\
                \bottomrule
            \end{tabularx}%
        }
    \caption{\new{Continual Learning Metrics }}
    \label{tab:cl_metrics_table}
\end{table}

\section{CL metrics for high-dimensional LASA}
\label{sec:app_lasa_highd}

\begin{table*}
  \captionsetup{justification=centering, labelsep=newline}
  \centering
  \caption{Hyperparameters used in our experiments. The same NODE and \snode{} architectures are used for SG, REP, FT, SI, MAS and CHN. Smaller networks for NODE and \snode{} are used for HN to keep the hypernetwork size comparable. Tangent vector scale is used only when orientations are learned for the RoboTasks9 dataset.}
  \label{tab:hparam}  
  \resizebox{0.95\textwidth}{!}{
      \input{tables_hparam.tex}
  }
\end{table*}

To compute CL metrics, DTW threshold values of 4000, 7000 and 15000 are set empirically for \lasa{8}, \lasa{16}, and \lasa{32} respectively. Then, for each of these high-dimensional LASA datasets, we compute CL metrics separately. Since each CL metric lies in the range [0,1], we report the average of the CL metrics over the 3 datasets in \mytable{tab:lasa_highd_cl}. The consolidated CL metrics of the 10 compared methods (MET) are shown.
Our proposed hypernetworks (\hnsnode{}, \chnsnode{}) achieve the highest CL scores among all compared methods. The use of \snode{} improves the CL performance of both \hnsnode{} and \chnsnode{}. \hnsnode{} is comparable to the upper baseline SG in terms of accuracy of predictions but learns all tasks with a single model.

\begin{figure*}
  \centering
  \includegraphics[width=0.7\textwidth]{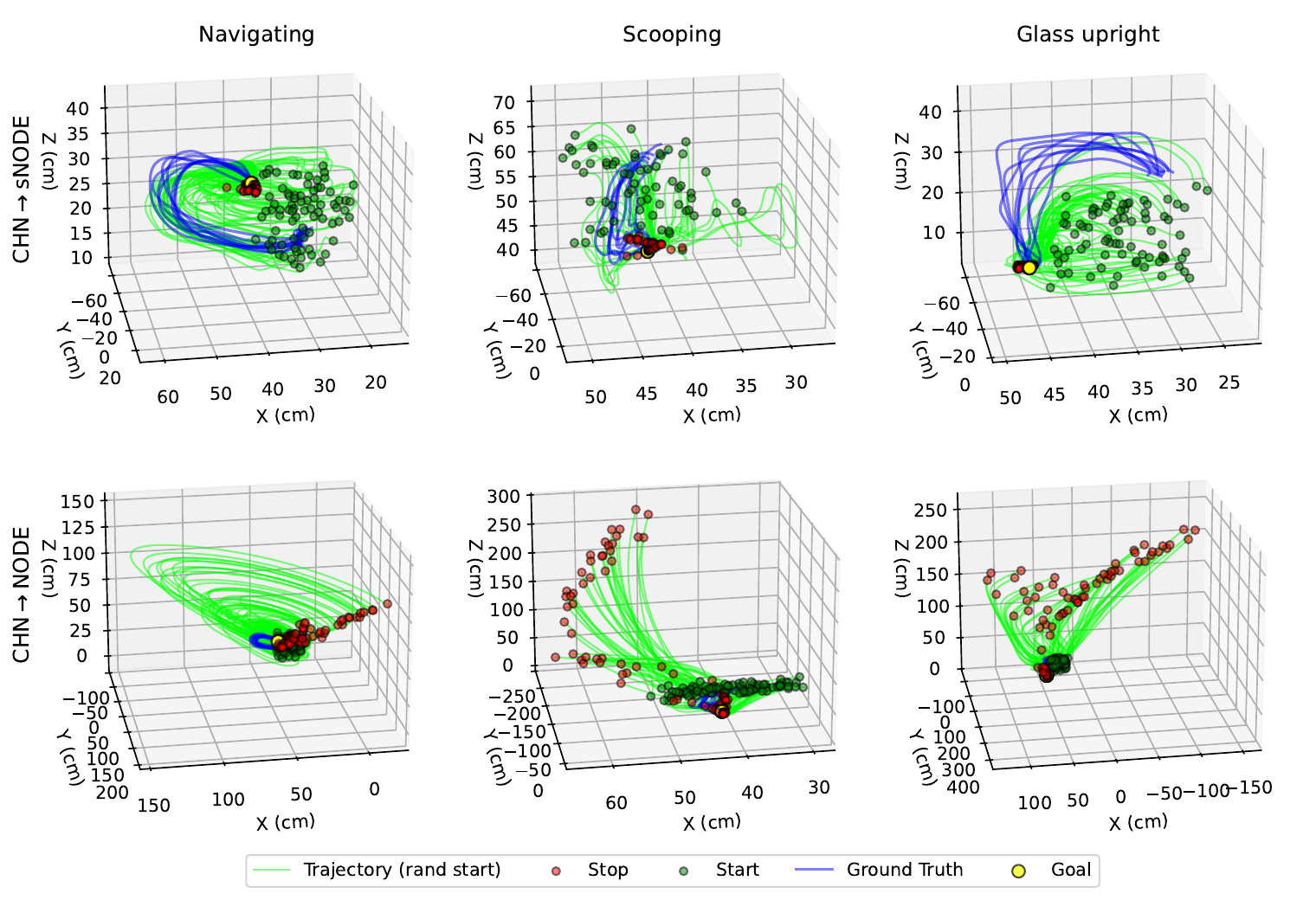}
  \caption[]{
    \new{Empirical stability test on the real-world \robot{}: when starting from random initial positions very different from the demonstrations, \chnsnode{} (top row) shows convergent behavior while \chnnode{} (bottom row) produces divergent trajectories.}  
  }
  \label{fig:stab_rt9_qual}    
\end{figure*}

\section{Additional Empirical Stability Tests}

\begin{figure}[t!]
  \centering
  \includegraphics[width=0.7\textwidth]{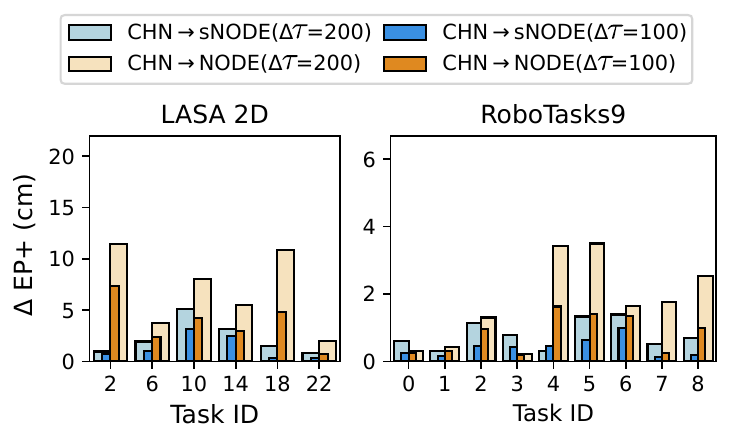}
  \caption[]{On the LASA 2D and RoboTasks9 datasets, \chnnode{} and \chnsnode{} are used to roll out trajectories of length 1100 ($\Delta\tau=100$) and 1200 ($\Delta\tau=200$) after being trained on 1000-step demonstrations. We measure \new{$\Delta_\text{EP+}$}, the distance between the demonstration goal and the end point of the predicted trajectories. For many tasks, \new{$\Delta_\text{EP+}$} for \chnsnode{} is lower than \chnnode{}.
  }
  \label{fig:stabtest_deltaextrasteps}
\end{figure}

In addition to the empirical stability tests presented in Sec.VI.F of the main paper, we also evaluate the stability of trajectory predictions for different rollout lengths. The demonstrations used for training the models are of 1000 steps. During inference, after the trajectory predicted by a model reaches the goal (or is sufficiently close to the goal), it should not deviate away from the goal. 
To test whether our predicted trajectories diverge from the goal after reaching it, we take the models trained on 1000-step demonstrations and roll out trajectories of 1100 and 1200 steps. We perform this evaluation on \lasa{2} and \robot{}, and repeat each evaluation 5 times with independent seeds. For each predicted trajectory, we measure \new{$\Delta_{\text{EP+}}$}, the distance of the final point of the predicted trajectory and the goal (i.e. the final point of the demonstration), where \new{$\Delta_{\text{EP+}}$ is defined as:}
\new{
\begin{equation}
    \Delta_\text{EP+} = \vert\vert \hat{\mathbf{p}}_{\tau-1} - \mathbf{p}_{T-1}\vert\vert_2 \label{eq:ep_plus}
\end{equation}
where $\mathbf{p}_{T-1}$ is the ground truth goal and $\hat{\mathbf{p}}_{\tau-1}$ is the predicted goal after $\tau > T$ steps.
}
\new{Both $\Delta_{\text{EP}}$~(Eq. 12 in the main paper) and $\Delta_{\text{EP}+}$~(\myequation{eq:ep_plus} in this supplementary document) measure the distance to the goal. However, for $\Delta_{\text{EP}}$, the initial position used for generating the predicted trajectory is different from the ground truth, and the lengths of the predicted and ground truth trajectories are the same. In contrast, for $\Delta_{\text{EP+}}$, the initial position stays unchanged, but the predicted trajectory is longer than the ground truth trajectory used to train on.}
As shown in \myfigure{fig:stabtest_deltaextrasteps}, $\Delta_\text{EP+}$ values for \chnnode{} for multiple tasks are higher than \chnsnode{} for 100 and 200 extra steps. Trajectories generated by \chnsnode{} remain close to the goal irrespective of the number of extra steps.

\new{
 \myfigure{fig:stab_rt9_qual} shows qualitative examples of trajectories generated by \chnsnode{} (top row) and \chnnode{}, when initial conditions are very different from the demonstrations. Please refer to Sec. VI.F in the main paper for the relevant discussion.
}

\newer{
\section{Learning from Noisy Demonstrations}
\label{sec:noisydemo}

With the help of \myfigure{fig:noisy_demo}, we present a short qualitative analysis showing that even noisy demonstrations can be used for training. We test standalone \snode{} models on noisy versions of the original demonstrations of a couple of tasks from the LASA 2D dataset. We added Gaussian noise to the trajectory points, followed by a some temporal smoothing. Even when trained on these noisy, imperfect demonstrations, the predictions were accurate and resembled the noise-free demonstrations. For these examples, this is achieved with the same number of training iterations as the other LASA 2D experiments and without applying any additional regularization or data denoising strategy. We defer a comprehensive quantitative analysis of this effect to future work, but would like to highlight that noise-robustness is an in-built feature of our proposed models since the vector field learned by the \snode{} implicitly induces smooth trajectories.
\begin{figure}[H]
    \centering
    \includegraphics[width=0.6\textwidth]{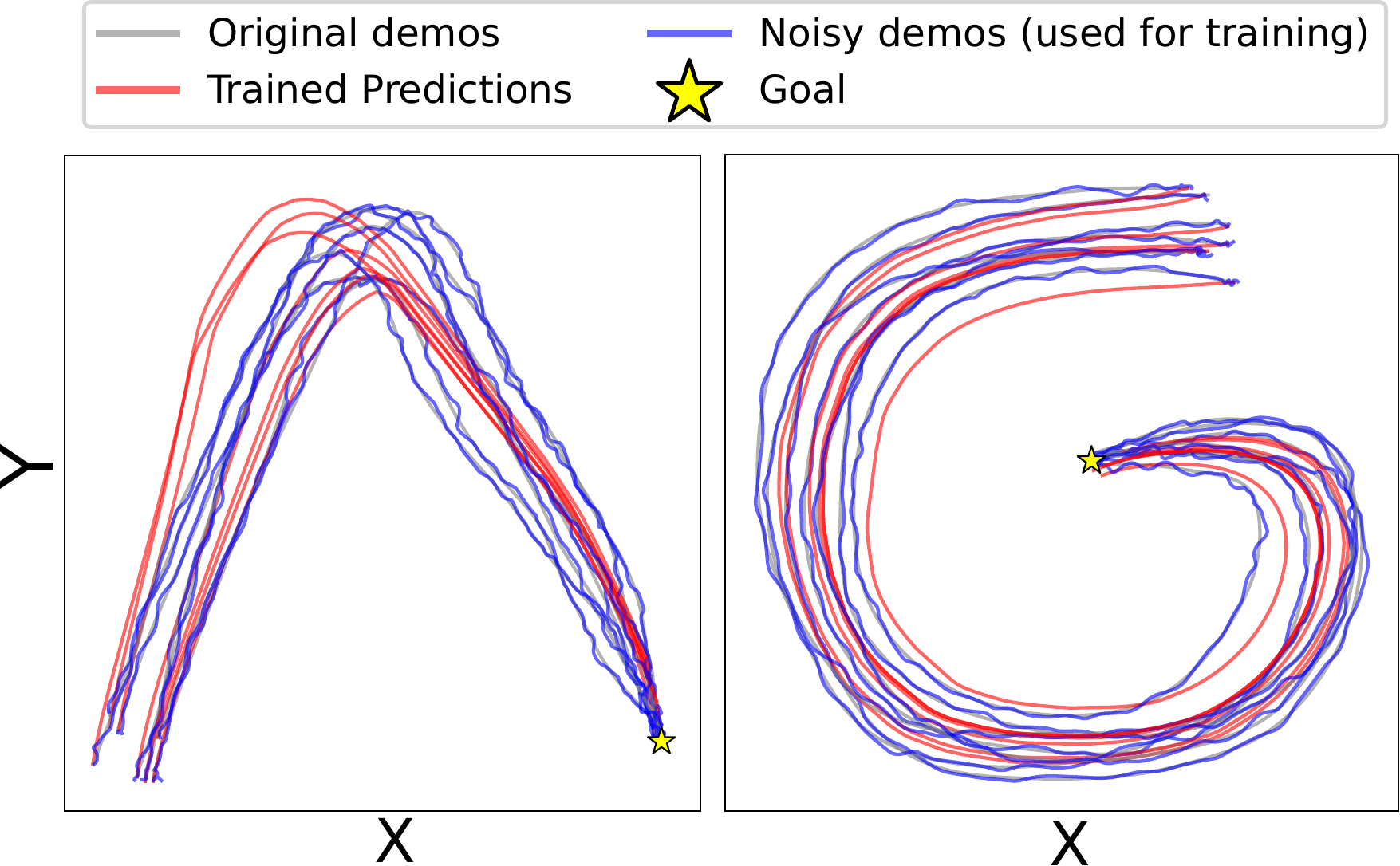}
    \caption{\newer{Qualitative examples of learning from noisy demonstrations.}}
    \label{fig:noisy_demo}
\end{figure}

}

\newer{
\section{Clock input's effect on accuracy}
\label{sec:clock_accu}

We introduce the clock input into the \snode{} to increase the accuracy of the predicted trajectories without affecting the stability of the model (our reasoning for the stability remaining unaffected is provided in Sec. IV.A of the main paper).

During training, the clock signal helps to disambiguate states that occur at trajectory intersections that are hard to learn considering only position-velocity pairs as they contain multimodal behaviors (same position, different velocity). For example, in the trajectories shown in \myfigure{fig:clock_accu}, the 2D position at the intersection shown by the blue circle needs to be traversed in two different directions shown by the black arrows. With the clock signal, we explicitly encode the phase of the trajectory (between 0 and 1) and thereby make it possible to predict the correct next state at the intersection. The augmented state comprising the original state and the clock signal always remains unique along a trajectory. Without the clock signal, the intersections lead to multimodal behaviors and become ambiguous and hard to learn.

A similar effect also occurs even for trajectories without intersections, such as trajectories that first move away from the goal and then come back to it (middle figure in \myfigure{fig:clock_accu}). Without the clock signal, there is no way for the system to know that it should not keep on approaching the goal the first time. This is also true when the start (green) and the goal (red) are close to each other (last figure in \myfigure{fig:clock_accu}). 

\begin{figure}[H]
    \centering
    \includegraphics[width=0.6\textwidth]{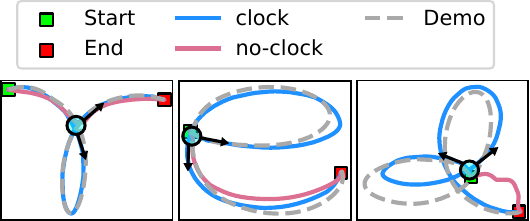}
    \caption{\newer{These examples from the HelloWorld dataset show trajectories where \snode{} with the clock signal helps to improve accuracy. The blue circle shows intersections and the black arrows show the possible directions of the next state.}}
    \label{fig:clock_accu}
\end{figure}
}

\begin{figure*}[b]
  \includegraphics[width=\textwidth]{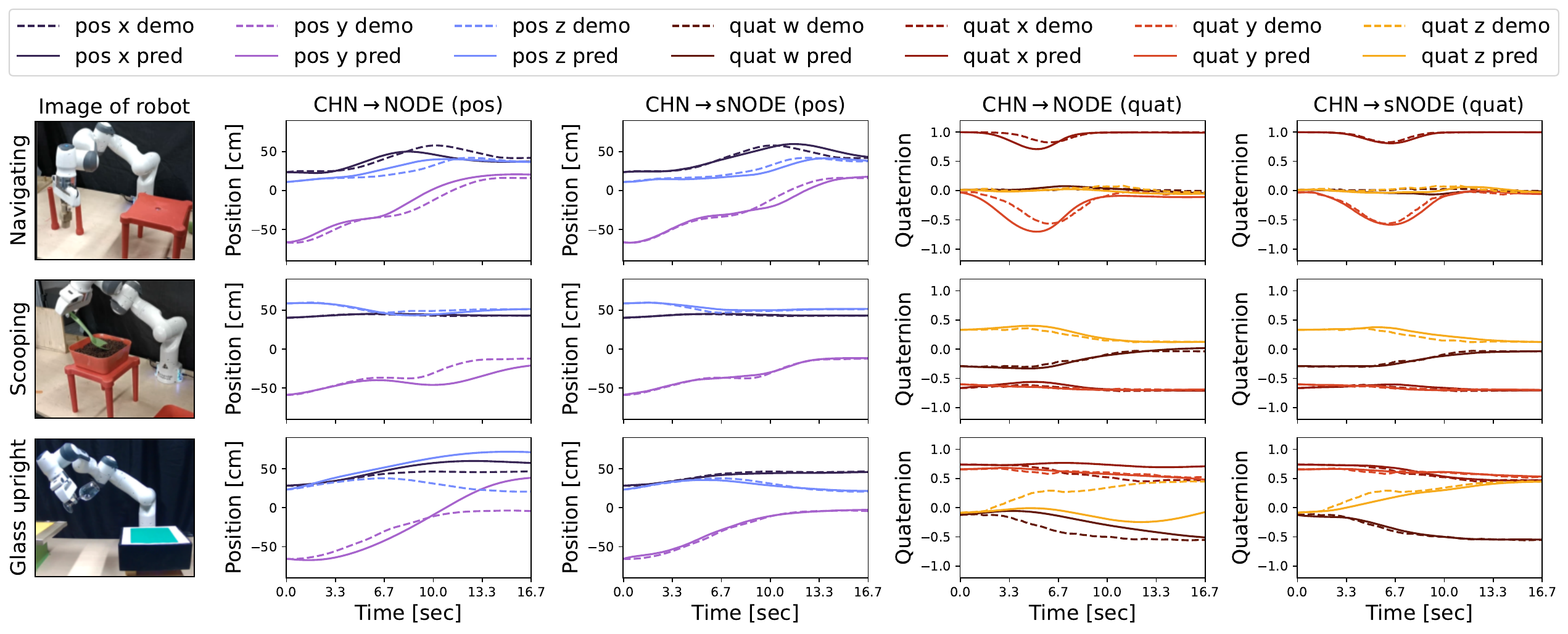}
  \caption{Qualitative examples from RoboTasks9. \chnnode{} and \chnsnode{} are trained sequentially on the 9 tasks of RoboTasks9. After all tasks are learned, each model performs all previous tasks.
  Examples of 3 tasks are shown, one in each row. 
  The first column shows the robot performing the task. The second and third columns show the positions achieved by \chnnode{} and \chnsnode{} respectively.
  The fourth and fifth columns show the orientations achieved by \chnnode{} and \chnsnode{} respectively. 
  Dotted lines denote demonstrations (\emph{demo}) and solid lines indicate predictions (\emph{pred}). 
  Note the larger errors of \chnnode{} compared to \chnsnode{} (difference between dotted and solid lines). 
  The robot can be seen performing the RoboTasks9 tasks in the supplementary video \url{https://youtu.be/qrESAnAk0U4}.
  }
  \label{fig:rt_qual}
\end{figure*}

\section{Hyperparameters}
\label{sec:hyperparameters}

\begin{figure}[H]
    \centering
    \includegraphics[width=0.9\textwidth]{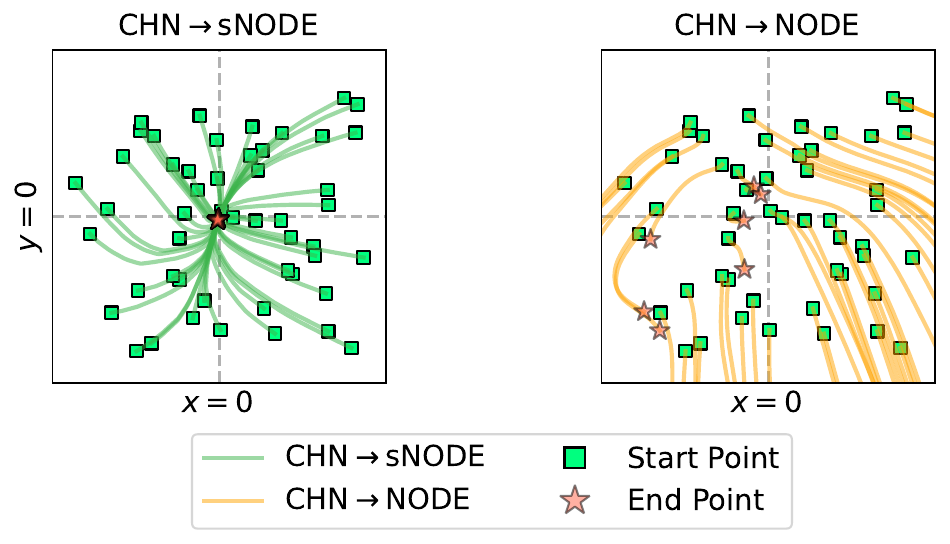}
    \caption{
    \new{Trajectories predicted by a randomly initialized \chnsnode{} (left) and \chnnode{} (right) for 50 randomly selected initial positions. The trajectories predicted by the untrained \chnsnode{} converge at the goal (origin), while the untrained \chnnode{} predicts random trajectories.}
    }
    \label{fig:untrained_chn}
\end{figure}

In \mytable{tab:hparam}, we present the complete set of hyperparameters used in our experiments. For experiments on \lasa{2} and \hw{}, we use the same hyperparameters as \cite{auddy2022continual} (\hw{} is only used for a non-CL experiment, see Sec. VI.A of the main paper). For \lasa{8} and \robot{}, we scale the number of training iterations roughly proportional to the dimension of the trajectories in the respective datasets. For \lasa{16} and \lasa{32}, we use fewer iterations than that suggested by this linear dimension-based scaling to keep the overall runtime of our experiments within acceptable bounds. Since \snode{} contains an extra network for the parameterized Lyapunov function, the architecture of the NODE and \snode{} models are adjusted (layers contain different number of neurons) so that the final parameter count of a NODE model and the corresponding \snode{} model is roughly equivalent.

\begin{figure}[H]
  \centering
  \includegraphics[width=0.75\textwidth]{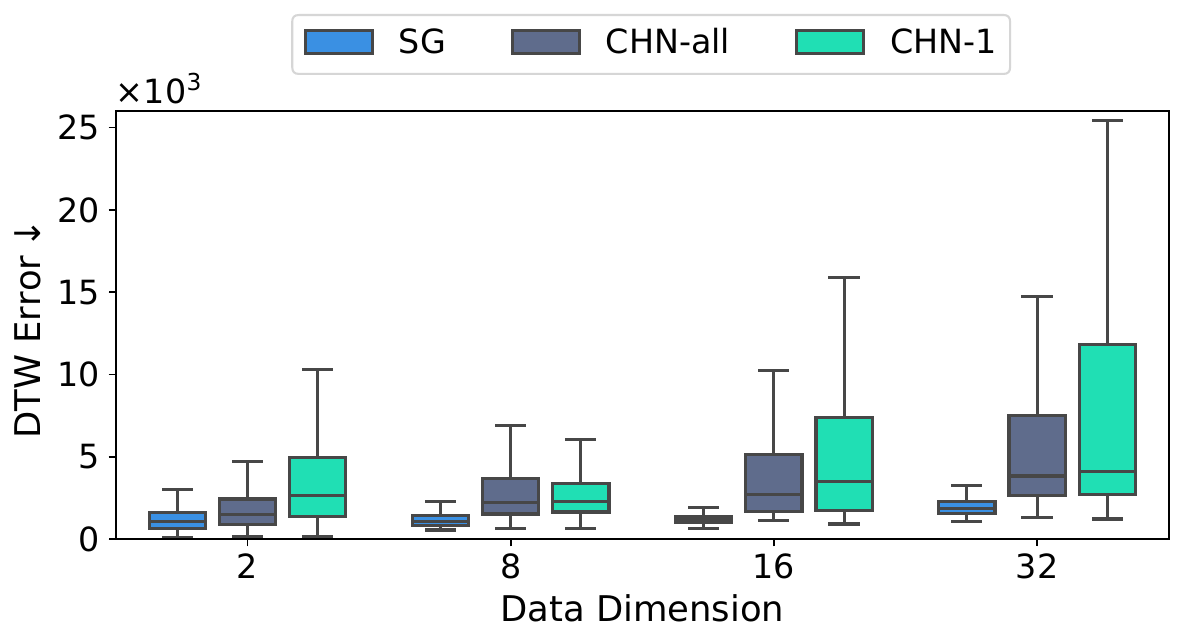}
  \caption{
      DTW errors (y-axis) for stochastic regularization in \chnsnode{} on LASA datasets of different dimensions (x-axis). Upper baseline SG (using \snode{}) has many more parameters than the CHN models. CHN-1 performs equivalently to CHN-all on LASA~8D, but shows higher variability than CHN-all in other cases.
  }    
  \label{fig:lasa_all_chn_stoch}
\end{figure}

\section{Additional Results}

\new{
In this section, we provide additional results. 
\myfigure{fig:lasa_all_chn_stoch} presents results for the comparison of the stochastic hypernetwork regularization on the LASA datasets of different dimensions (please refer to Sec.VI.E of the main paper).

\myfigure{fig:rt_qual} presents qualitative results for experiments performed on the real-world \robot{} dataset (please refer to Sec.V1.D of the main paper).

Finally, \myfigure{fig:untrained_chn} presents qualitative results to show that even an untrained \chnsnode{} produces non-divergent predictions (please refer to the discussion in Sec.VII of the main paper).
}


\end{document}

%% file: figures/LASA_highD_all.tex
\begin{tabular}{lrrrrrrrr}
\toprule
MET & ACC & REM & MS & TE & FS & SSS & CL$_{sco}$ & CL$_{stab}$ \\
\midrule
SG [N] & 0.99 & \cellcolor{white} \bfseries 1.00 & 0.29 & \cellcolor{white} \bfseries 1.00 & 0.00 & \cellcolor{white} \bfseries 1.00 & 0.71 & 0.55 \\
FT [N] & 0.18 & 0.02 & 1.00 & 0.99 & \cellcolor{white} \bfseries 0.89 & \cellcolor{white} \bfseries 1.00 & 0.68 & 0.55 \\
REP [N] & 0.74 & 0.99 & 1.00 & 0.99 & \cellcolor{white} \bfseries 0.89 & 0.45 & 0.85 & 0.78 \\
HN${\scriptstyle \rightarrow}$N & 0.92 & 0.99 & \cellcolor{white} \bfseries 1.00 & 0.64 & 0.65 & \cellcolor{white} \bfseries 1.00 & 0.87 & 0.82 \\
CHN${\scriptstyle \rightarrow}$N & 0.72 & 0.99 & 1.00 & 0.60 & 0.86 & \cellcolor{white} \bfseries 1.00 & 0.86 & 0.82 \\
SG [sN] & \cellcolor{white} \bfseries 1.00 & \cellcolor{white} \bfseries 1.00 & 0.29 & 1.00 & 0.00 & \cellcolor{white} \bfseries 1.00 & 0.71 & 0.55 \\
FT [sN] & 0.18 & 0.02 & 1.00 & 0.99 & 0.87 & \cellcolor{white} \bfseries 1.00 & 0.68 & 0.55 \\
REP [sN] & 0.77 & \cellcolor{white} \bfseries 1.00 & 1.00 & 0.99 & 0.87 & 0.45 & 0.85 & 0.78 \\
HN${\scriptstyle \rightarrow}$sN & 0.97 & 1.00 & 1.00 & 0.85 & 0.64 & \cellcolor{white} \bfseries 1.00 & \underline{0.91} & \underline{0.85} \\
CHN${\scriptstyle \rightarrow}$sN & 0.86 & 1.00 & 1.00 & 0.81 & 0.84 & \cellcolor{white} \bfseries 1.00 & \cellcolor{white} \bfseries 0.92 & \cellcolor{white} \bfseries 0.90 \\
\bottomrule
\end{tabular}

%% file: figures/tables_hparam.tex
\begin{tabular}{l|llllll}
\midrule
Dataset & LASA 2D & HelloWorld & LASA 8D & LASA 16D & LASA 32D & RoboTasks9 \\
Dimension & 2 & 2 & 8 & 16 & 32 & 6 \\
No. of tasks & 26 & 7 & 10 & 10 & 10 & 9 \\  
Iterations & $1.5\times 10^{4}$ & $4.0\times 10^{4}$ & $6.0\times 10^{4}$ & $7.0\times 10^{4}$ & $8.0\times 10^{4}$ & $4.0\times 10^{4}$ \\
LR & $1.0\times 10^{-4}$ & $1.0\times 10^{-4}$ & $5.0\times 10^{-5}$ & $5.0\times 10^{-5}$ & $5.0\times 10^{-5}$ & $5.0\times 10^{-5}$ \\
SI C & 0.3 & - & - & - & - & - \\
SI $\epsilon$ & 0.3 & - & - & - & - & - \\
MAS $\lambda$ & 0.1 & - & - & - & - & - \\
arch(NODE) & [1000]x2, 1015 & [1000]x2, 1015  & [1000]x3, 1015 & [1000]x4, 1015 & [1000]x5, 1015 & [1000]x2, 1015 \\
arch(\snode{} $\hat{\mathbf{f}}$) & [1000]x3  & [1000]x3  & [1000]x3 & [1000]x3 & [1000]x3 & [1000]x3 \\
arch(\snode{} $V$) & [100]x2  & [100]x2  & [100]x2 & [100]x2 & [100]x2 & [100]x2 \\
arch(HN NODE) & [100]x2, 150  & -  & [100]x3, 150 & [100]x4, 150 & [80]x5, 100 & [100]x2, 150 \\
arch(HN \snode{} $\hat{\mathbf{f}}$)  & [100]x3 & - & [90]x3 & [75]x3 & [60]x3 & [100]x3 \\
arch(HN/CHN) & [200]x3  & -  & [300]x3 & [300]x3 & [350]x3 & [300]x3 \\
dim(Task Emb) & 256  & -  & 256 & 256 & 512 & 256 \\
HN/CHN $\beta$ & $5.0\times 10^{-3}$  & -  & $5.0\times 10^{-3}$ & $5.0\times 10^{-3}$ & $5.0\times 10^{-3}$ & $5.0\times 10^{-3}$ \\
dim(CHN Chunk Emb) & 256  & -  & 256 & 256 & 512 & 256 \\
dim(CHN Chunk) & 8192  & -  & 16384 & 16384 & 16384 & 8192 \\
Tangent vector scale & -  & -  & - & - & - & 5.0 \\
\bottomrule
\end{tabular}